\documentclass[12pt,a4paper]{article}

\usepackage[utf8]{inputenc}
\usepackage[english]{babel}
\usepackage[margin=1in]{geometry}
\usepackage{graphicx}
\usepackage{float}
\usepackage{amsthm}
\usepackage{amsmath}
\usepackage{mathtools}
\usepackage{amsfonts}
\usepackage{amssymb}
\usepackage{bbm}
\usepackage{bm}
\usepackage{tikz}
\usepackage{algpseudocode}
\usepackage{breqn}
\usepackage[hyphens]{url}
\usepackage{nicefrac}
\usepackage[multiple]{footmisc}
 \usepackage{hyperref}
\usepackage{enumitem}
\usepackage{thmtools}
\usepackage{thm-restate}
\usepackage[nameinlink, capitalize]{cleveref}

\usepackage{cprotect}
\usepackage{etoc}
\usepackage{titlesec}
\titleformat{\section}
  {\normalfont\fontsize{16}{16}\bfseries}{\thesection}{1em}{}
\titleformat{\subsection}
  {\normalfont\fontsize{14}{16}\bfseries}{\thesubsection}{1em}{}
\usepackage{minitoc}
\usepackage{etaremune}

\usepackage[T1]{fontenc}    
\usepackage{booktabs}       
\usepackage{microtype}      
\floatstyle{ruled}
\usepackage{xcolor}

\usepackage{caption}
\usepackage{subcaption}

\usetikzlibrary{arrows,automata, calc, positioning}
\usepackage{adjustbox}
\usepackage{wrapfig}

\usepackage{multirow}
\usepackage{accents}

\usepackage{etoolbox}
\usepackage{xpatch}

\DeclareMathOperator*{\argmin}{\arg\!\min}

\theoremstyle{definition}

\newtheorem{theorem}{Theorem}
\newtheorem{proposition}{Proposition}
\newtheorem{definition}{Definition}
\newtheorem{lemma}{Lemma}

\newtheorem{corollary}{Corollary}
\newtheorem{assumption}{Assumption}
\newtheorem{remark}{Remark}
\newtheorem{example}{Example}

\let\originalleft\left
\let\originalright\right
\renewcommand{\left}{\mathopen{}\mathclose\bgroup\originalleft}
\renewcommand{\right}{\aftergroup\egroup\originalright}
\DeclareRobustCommand{\VAN}[2]{#1}  

\begin{document}

\def\spacingset#1{\renewcommand{\baselinestretch}%
{#1}\small\normalsize} \spacingset{1}


  \title{Kernel Ridge Riesz Representers: \\
Generalization, Mis-specification, and the Counterfactual Effective Dimension}
  \author{
  Rahul Singh\thanks{Email: \url{rahul_singh@fas.harvard.edu}. Address: Littaeur Center 123, 1805 Cambridge Street, Cambridge, MA 02138. I thank Alberto Abadie, David Bruns-Smith, Victor Chernozhukov, Avi Feller, Arthur Gretton, Anna Mikusheva, Whitney Newey, Vasilis Syrgkanis, Suhas Vijaykumar, and Liyuan Xu for helpful discussions. I thank William Liu and Moses Stewart for excellent research assistance. I received support from the Hausman Dissertation Fellowship. Part of this work was done while visiting the Simons Institute for the Theory of Computing.
  } \\
  Harvard University}
  \date{Original draft: February 2021. This draft: June 2024.}
\maketitle
%
\begin{abstract}
Kernel balancing weights provide confidence intervals for average treatment effects, based on the idea of balancing covariates for the treated group and untreated group in feature space, often with ridge regularization. Previous works on the classical kernel ridge balancing weights have certain limitations: 
(i) not articulating generalization error for the balancing weights, 
(ii) typically requiring correct specification of features, and
(iii) justifying Gaussian approximation for only average effects.

I interpret kernel balancing weights as kernel ridge Riesz representers (KRRR) and address these limitations via a new characterization of the counterfactual effective dimension. KRRR is an exact generalization of kernel ridge regression and kernel ridge balancing weights. I prove strong properties similar to kernel ridge regression: population $L_2$ rates controlling generalization error, and a standalone closed form solution that can interpolate. The framework relaxes the stringent assumption that the underlying regression model is correctly specified by the features. It extends Gaussian approximation beyond average effects to heterogeneous effects, justifying confidence sets for causal functions. 
I use KRRR to quantify uncertainty for heterogeneous treatment effects, by age, of 401(k) eligibility on assets.
\end{abstract}

\noindent%
\vfill


\newpage

\spacingset{1.65} 

\section{Introduction}\label{sec:intro}

Kernel balancing weights adapt kernel methods to quantify uncertainty on average effects, e.g. the average treatment effect when the treatment is randomly assigned conditional on covariates.\footnote{Kernel ridge balancing weights also apply to confidence intervals for the average treatment effects on the treated and untreated subpopulations, as well as subpopulations defined in terms of a discrete covariate taking a finite number of values. This paper justifies confidence sets for causal functions.} 
The central idea is to balance the feature-mapped covariates in the treated group with those in the untreated group, subject to regularization. The weights that achieve such balance, when applied to the observed outcomes, estimate the average treatment effect and appear in its confidence interval. These weights are widely used in social science via popular \texttt{R} packages, and they are classical in statistics \cite{speckman1979minimax}.

The motivation for kernel balancing weights is to inherit the simplicity and flexibility of kernel methods. However, not all of the familiar properties of, say, kernel ridge regression have been fully articulated for classical kernel ridge balancing weights, e.g. population $L_2$ rates of convergence for generalization error, a simple standalone closed form solution, and robustness to mis-specification of features. Such properties would lead to strong semiparametric guarantees for causal parameters via well-known ``targeting'' and ``debiasing'' arguments, which tolerate some types of mis-specification and which provide Gaussian approximation for causal functions, justifying confidence sets with nominal coverage.

This paper's main contribution is a population $L_2$ rate controlling the generalization error of the classical kernel ridge balancing weights. The rate coincides with the lower bound of regression in the reproducing kernel Hilbert space (RKHS) framework. It holds under mis-specification of features. A technical innovation powers the result: I define the counterfactual effective dimension, and prove it is not much greater than the actual effective dimension. This technique may be of independent interest.

As a secondary contribution, this paper provides a standalone closed form solution for kernel ridge balancing weights that appears to have been previously unknown. This standalone closed form solution can be evaluated at locations different than the training data, allowing kernel ridge balancing weights to be used in ``debiased'' treatment effect estimators with sample splitting. Such estimators accommodate regression models outside of the RKHS, unlike many previous works on kernel ridge balancing weights.

Conceptually, I study the kernel ridge Riesz representer (KRRR) as an exact generalization of kernel ridge regression and kernel ridge balancing weights. This generalization formalizes the sense in which the population $L_2$ rate is sometimes optimal. For treatment effects, I verify inference with rate-optimal regularization, rather than undersmoothing, of the ridge penalty. 
The generalization also justifies the construction of confidence sets for causal functions, e.g. heterogeneous policy effects and heterogeneous treatment effects, from kernel ridge balancing weights.\footnote{Heterogeneous treatment effects are a function of a continuous covariate; see Section~\ref{sec:experiments}.}

Section~\ref{sec:related} situates these contributions within the context of related work. Section~\ref{sec:problem} formalizes the key assumptions. Section~\ref{sec:algorithm} describes the procedure, derives its standalone closed form, and clarifies how it extends known procedures. 
Section~\ref{sec:mse} theoretically justifies the procedure with population $L_2$ rates of generalization error, which imply strong semiparametric guarantees.
Section~\ref{sec:sketch} proves the rates via the counterfactual effective dimension.
Section~\ref{sec:experiments} demonstrates nominal coverage in heterogeneous treatment effect simulations, with less bias than some previous estimators, then quantifies uncertainty for the heterogeneous treatment effects of 401(k) eligibility on assets as a function of age. The effect appears to be five times stronger for 45-60 year olds, compared to 30 year olds.

\section{Related work}\label{sec:related}

I extend the conceptual framework of kernel balancing weights \cite{speckman1979minimax,hazlett2020kernel,wong2018kernel,zhao2019covariate,kallus2020generalized,hirshberg2019augmented,hirshberg2019kernel,chernozhukov2020adversarial}.  
Works in this literature often focus on kernel ridge balancing weights as a vector that possibly converges in $\ell_2$ \cite{hirshberg2019augmented,hirshberg2019kernel}, whereas I will study them as evaluations of a function that converges in population $L_2$. 
The former is the training error for a balancing weight that is defined in-sample; see Corollary~\ref{cor:hirshberg}. 
By contrast, I analyze generalization error for a balancing weight that can interpolate and that can be evaluated out-of-sample. 
This distinction is important for stronger semiparametric conclusions: coverage guarantees for causal functions, sample splitting, and regression models beyond the RKHS; see Remark~\ref{remark:spec}. 

\cite{bruns2023augmented} prove equivalences among various formulations of the classical kernel ridge balancing weights, including a minimax formulation with ridge regularization. 
Regardless of nomenclature, many of the salient properties of these classical balancing weights, such as population $L_2$ rates of generalization error and a standalone closed form, appear to be contributions.
I recover
the known, nonasymptotic balance-variance trade-off. 

The properties that I prove for KRRR generalize known results for kernel ridge regression. I obtain $L_2$ rates using integral operator techniques for kernel methods, drawing on \cite{smale2007learning,caponnetto2007optimal,fischer2017sobolev}, among others. See \cite{fischer2017sobolev} for references and a review of integral operator techniques and empirical process techniques used to analyze RKHS estimators in various norms, many of which rely on the effective dimension. My characterization of the counterfactual effective dimension appears to be new, and it avoids auxiliary approximation assumptions. The closed form solution of KRRR echoes classical solutions  \cite{kimeldorf1971some,scholkopf2001generalized}.

I also build on the insight from semiparametric theory that the balancing weights for the average treatment effect are a special case of Riesz representers for causal parameters, which can be directly estimated \cite{robins2007comment,avagyan2021high,chernozhukov2018global,hirshberg2019augmented,chernozhukov2018learning,smucler2019unifying,hirshberg2019kernel,chernozhukov2020adversarial}. Similar to \cite{chernozhukov2018global,chernozhukov2018learning,chernozhukov2020adversarial}, I provide population $L_2$ rates. A key point of departure is that I study KRRR, which is a different estimator more closely aligned with kernel ridge regression, classical kernel ridge balancing weights, and popular \texttt{R} packages. Another distinction is that I place standard RKHS assumptions, which facilitates comparison with known optimality theory, including rate-optimal regularization.
Finally, I avoid some auxiliary approximation assumptions of those works by defining and analyzing the counterfactual effective dimension; see Remark~\ref{remark:aux}. 

The contribution of this work is not to provide new semiparametric theory. Rather, I demonstrate that KRRR has good properties similar to kernel ridge regression that were not fully articulated, and hence it is compatible with strong existing results, e.g. \cite{zheng2011cross,chernozhukov2018original,rotnitzky2021characterization,kennedy2023towards}, among many others. These strong results allow for some mis-specification and rate-optimal regularization, rather than the correct specification and undersmoothing studied by e.g. \cite{hirshberg2019kernel,mou2023kernel}. See e.g. \cite{hirshberg2019augmented,ben2021balancing} for detailed citations on balancing weights outside of the RKHS. See e.g. \cite{van2011targeted} for detailed citations on semiparametrics.

An initial draft was circulated in February 2021 \cite{singh2021debiased}. Prior to the current draft, \cite{wang2021estimation} extend the kernel balancing weights of \cite{wong2018kernel} to heterogeneous treatment effects and prove consistency under correct specification, without formal inference, a closed form, or $L_2$ rates for balancing weights. This paper's results are complementary: formal inference, some mis-specification, a closed form, and $L_2$ rates for balancing weights, within a framework that includes other estimands.\footnote{By formal inference, I mean justification for Gaussian approximation and nominal coverage guarantees.} Subsequent works extend kernel balancing weights to confounding bridges \cite{kallus2021causal,ghassami2021minimax} and study other nonlinear function spaces \cite{chernozhukov2021automatic}.
\section{Key assumptions and main results}\label{sec:problem}

\subsection{Warm up: Balancing weights}

Consider the problem of estimating the average treatment effect (ATE) $\theta_0$ from $n$ independently and identicically distributed (i.i.d.) observations of the baseline covariates $X\in\mathbb{R}^{dim(X)}$, treatment $D\in\{0,1\}$, and outcome $Y \in\mathbb{R}$. When treatment assignment is as good as random conditional on $X$, the ATE can be expressed in two ways. First, the regression formulation is $\theta_0=\mathbb{E}\{\gamma_0(1,X)-\gamma_0(0,X)\}$, where $\gamma_0(D,X)=\mathbb{E}(Y|D,X)$ is the outcome regression. Second, the balancing weight formulation is $\theta_0=\mathbb{E}\{Y\alpha_0(D,X)\}$, where $\alpha_0(D,X)=D\pi_0(X)^{-1}-(1-D)\{1-\pi_0(X)\}^{-1}$ is the balancing weight expressed as a function and $\pi_0(X)=\mathbb{E}(D|X)$ is the propensity score. One may derive the latter from the former using the law of iterated expectations and the regularity condition that the propensity score $\pi_0$ is bounded away from zero and one.

These two formulations suggest estimation approaches. One option is to estimate the regression as $\hat{\gamma}$ then the ATE as $\mathbb{E}_n\{\hat{\gamma}(1,X)-\hat{\gamma}(0,X)\}$, where $\mathbb{E}_n(\cdot)$ is the empirical mean. Another is to estimate the balancing weight as $\hat{\alpha}$ then the ATE as $\mathbb{E}_n\{Y\hat{\alpha}(D,X)\}$. One may also combine these estimators into a so-called doubly robust procedure $\mathbb{E}_n[\hat{\gamma}(1,X)-\hat{\gamma}(0,X)+\hat{\alpha}(D,X)\{Y-\hat{\gamma}(D,X)\}]$.  
Each approach also has a sample splitting variant, 
where $(\hat{\gamma},\hat{\alpha})$ are estimated from, say $n/2$ held out observations, and the empirical mean is taken with respect to the remaining $n/2$ observations. For the sample splitting variant, $\hat{\alpha}$ must be a function that can be evaluated out-of-sample.

The sample splitting variant of the doubly robust procedure is sometimes called ``debiased'' machine learning \cite{chernozhukov2018original}, which is closely related to targeted learning \cite{van2006targeted} and one step estimation \cite{pfanzagl1982lecture}, and it has quite general inferential guarantees for $\theta_0$. In particular, it allows for mis-specification of $\hat{\gamma}$ or $\hat{\alpha}$, and rate-optimal regularization of $\hat{\gamma}$ and $\hat{\alpha}$, as long as population $L_2$ rates for $\hat{\gamma}$ and $\hat{\alpha}$ are known, i.e. formal characterizations of $\|\hat{\gamma}-\gamma_0\|_2$ and $\|\hat{\alpha}-\alpha_0\|_2$ where $\|W\|_2=\{\mathbb{E}(W^2)\}^{1/2}$.

Previous work proposes a kernel ridge balancing weight $\tilde{\alpha}$. My main contribution is to analyze its generalization error $\|\tilde{\alpha}-\alpha_0\|_2$, achieving the well-known lower bound for regression in the RKHS framework, without auxiliary approximation assumptions. Once a population $L_2$ rate is articulated, general semiparametric guarantees from the literature become available. These guarantees justify Gaussian approximation for causal functions and some mis-specification, unlike many previous works on kernel ridge balancing weights.

\subsection{Generalizing from balancing weights to Riesz representers}

The ATE is a special case of a causal parameter. This paper considers parameters of the form $\theta_0=\mathbb{E}\{m(W,\gamma_0)\}$, where $W$ concatenates observed variables, $\gamma_0(W)=\mathbb{E}(Y|W)$ is a regression, and the formula $m$ accommodates various parameters beyond the ATE. For ATE, $W=(D,X)$ and $m(w,f)=f(1,x)-f(0,x)$.
I study a class defined by a regularity condition generalizing the idea that the propensity score is away from zero and one.

\begin{assumption}[Mean square continuity]\label{assumption:cont}
    The functional $f\mapsto \mathbb{E}\{m(W,f)\}$ is linear. There exists some $\bar{M}<\infty$ such that $\mathbb{E}\{m(W,f)^2\}\leq \bar{M} \|f\|_2^2$ for all $f \in L_2$.
\end{assumption}

Intuitively, this condition imposes that if $f_1$ and $f_2$ are close, then their functionals $\mathbb{E}\{m(W,f_1)\}$ and $\mathbb{E}\{m(W,f_2)\}$ are close. In the ATE example, if $\pi_0(X)\in(1-\bar{\pi},\bar{\pi})$ almost surely, then $\bar{M}=2\{\bar{\pi}^{-1}+(1-\bar{\pi})^{-1}\}$. 

Nonasymptotic analysis allows $\bar{M}$ to be a diverging sequence as the sample size increases. In particular, this class includes functionals of the form $f\mapsto \mathbb{E}\{m_h(W,f)\}$ where $m_h(w,f)=\ell_h(w)\check{m}(w,f)$. The weighting $\ell_h(w)$ is indexed by a vanishing bandwidth $h\downarrow 0$, which also indexes $\bar{M}=\bar{M}_h\uparrow \infty$. In this thought experiment, a sequence of semiparametric quantities, each satisfying Assumption~\ref{assumption:cont}, provides a pointwise approximation to a nonparametric causal function evaluated at a point. 
I provide two concrete examples.

\begin{example}[Heterogeneous policy effects]\label{ex:policy}
    Here $W=X$ are regressors and $\check{m}(w,f)=f\{t(x)\}-f(x)$, where $t$ is a known mapping that serves as the counterfactual policy of interest.  
    The weighting $\ell_h(w)=\textsc{local}\{(x_1-x_1^*)/h\}$ is a localization around the location $X_1=x_1^*$ (see Appendix~\ref{sec:nonparametric} for details). Suppose that the density ratio is bounded above, i.e. $\omega(W)=\textsc{density}\{t(W)\}/\textsc{density}(W)\leq\bar{\omega}$ almost surely. Suppose that the weighting function at a fixed $h$ is bounded, i.e. $|\ell_h(W)|\leq \bar{\ell}_h$ almost surely. Then $\bar{M}_h=2\bar{\ell}_h^2(\bar{\omega}+1)$.
\end{example}

\begin{example}[Heterogeneous treatment effects]\label{ex:het}
Here $W=(D,X)$ concatenates the treatment $D\in\{0,1\}$ and covariates $X$, and $\check{m}(w,f)=f(1,x)-f(0,x)$. The weighting $\ell_h(w)=\textsc{local}\{(x_1-x_1^*)/h\}$
is a localization around the location $X_1=x_1^*$ (see Appendix~\ref{sec:nonparametric} for details). Suppose that the propensity score is bounded away from zero and one, i.e. $\pi_0(X)\in(1-\bar{\pi},\bar{\pi})$ almost surely. Suppose that the weighting function at a fixed $h$ is bounded, i.e. $|\ell_h(W)|\leq \bar{\ell}_h$ almost surely. Then $\bar{M}_h=2\bar{\ell}_h^2\{\bar{\pi}^{-1}+(1-\bar{\pi})^{-1}\}$.
\end{example}

For further interpretation, suppose the the first covariate $X_1$ is continuous. In the finite sample, Example~\ref{ex:het} is the heterogeneous treatment effect for the subpopulation whose first covariate is local to $x_1^*$. As $h\downarrow 0$, Example~\ref{ex:het} converges to the heterogeneous treatment effect for the subpopulation whose first covariate value is $x_1^*$. 
Section~\ref{sec:experiments} gives a real world example where formal uncertainty quantification leads to economic insights.

Taking $\ell_h(w)=1$, Examples~\ref{ex:policy} and~\ref{ex:het} reduce to the average policy effect and average treatment effect, respectively. The class defined by Assumption~\ref{assumption:cont} contains not only average effects but also heterogeneous effects, i.e. pointwise approximations of causal functions.

The balancing weight is a special case of a Riesz representer. It is well known that each parameter within the class defined by Assumption~\ref{assumption:cont} has both a regression formulation and a generalized balancing weight formulation.

\begin{lemma}[Riesz representation; Lemma S3.1 of \cite{chernozhukov2018global}]\label{lemma:riesz}
    Suppose Assumption~\ref{assumption:cont} holds and $\gamma_0\in \mathcal{G} \subset L_2$. Then there exists a Riesz representer $\alpha_0\in L_2$ such that 
$
\mathbb{E}\{m(W,f)\}=\mathbb{E}\{\alpha_0(W)f(W)\}$ for all $f\in \mathcal{G}$. 
Moreover, there exists a unique minimal Riesz representer $\alpha_0^{\min}\in closure\{span(\mathcal{G})\}$ that satisfies this equation.
\end{lemma}

For the rest of the paper, I will set aside the ``correct specification'' assumption that $\gamma_0\in \mathcal{G}$, where $\mathcal{G}$ is a known subset of $L_2$ that can be imposed in estimation of $\hat{\gamma}$. As such, there exists a unique Riesz representer $\alpha_0=\alpha_0^{\min}$. See Remark~\ref{remark:spec} for a list of many works that place ``correct specification'' style assumptions requiring $\gamma_0$ to be within the RKHS.

For a functional $f\mapsto \mathbb{E}\{m_h(W,f)\}$ where $m_h(w,f)=\ell_h(w)\check{m}(w,f)$, its Riesz representer is $\alpha_h(w)=\ell_h(w)\check{\alpha}_0(w)$, where $\check{\alpha}_0(w)$ is the Riesz representer for $f\mapsto \mathbb{E}\{\check{m}(W,f)\}$.

In summary, previous works on classical kernel ridge balancing weights provide formal inference for a limited class of causal parameters, i.e. average effects. They also typically require correct specification of $\gamma_0$ within the RKHS. I study the broader class defined in Assumption~\ref{assumption:cont}, which includes heterogeneous policy effects (Example~\ref{ex:policy}) and heterogeneous treatment effects (Example~\ref{ex:het}). I take the agnostic stance of $\gamma_0\in L_2$. By analyzing generalization error $\|\tilde{\alpha}-\alpha_0\|_2$, general semiparametric guarantees become immediate.

\subsection{RKHS notation and key assumptions}\label{sec:problem_rkhs}

The kernel ridge balancing weight algorithm conducts estimation in a  reproducing kernel Hilbert space (RKHS) $H\subset L_2$. $H$ consists of functions of the form $f:\mathcal{W}\rightarrow\mathbb{R}$. I denote its symmetric, positive definite kernel by $k:\mathcal{W}\times\mathcal{W}\rightarrow \mathbb{R}$, and its feature map by $\phi:w\mapsto k(w,\cdot)$, so that $k(w,w')=\langle \phi(w),\phi(w') \rangle_H$ and $f(w)=\langle f,\phi(w) \rangle_H$ for $f\in H$.\footnote{Assumptions~\ref{assumption:original} and~\ref{assumption:RKHS} below formalize weak regularity conditions.} 

I use familiar spectral notation from kernel ridge regression analysis. Let $L:L_2\rightarrow L_2$ be the convolution operator $f\mapsto \int k(\cdot,w) f(w) \mathrm{d}\mathbb{P}(w)$. Since $L$ is self adjoint and positive definite, it has weakly decreasing countable eigenvalues $(\eta_j)$ and corresponding eigenfunctions $(\varphi_j)$. Define the space $
  H^c=(f=\sum_{j=1}^{\infty}f_j\varphi_j:\sum_{j=1}^{\infty} f_j^2\eta_j^{-c} <\infty)
$. In this notation, $H^0=L_2$ and $H^1=H$ when $k$ is characteristic \cite{sriperumbudur2010relation}. The RKHS $H$ is the subset of $L_2$ for which higher order terms in the series $(\varphi_j)$ have a smaller contribution.  By Mercer's theorem, the feature map is
$
\phi(w)=\{\eta_j^{1/2} \varphi_j(w)\}^{\infty}_{j=1}$.

I also use the extended notation of \cite{chernozhukov2020adversarial}, whose critical insight is that causal inference introduces counterfactual features. Since each $\varphi_j$ is in $L_2$, $m(w,\varphi_j)$ is well defined. I write the counterfactual feature map as $\phi^{(m)}(w)=\{\eta_j^{1/2} m(w,\varphi_j)\}^{\infty}_{j=1}$. The counterfactual kernel is then $k^{(m)}(w,w')=\langle \phi^{(m)}(w),\phi^{(m)}(w') \rangle_H$, and the counterfactual evaluation is $m(w,f)=\langle f, \phi^{(m)}(w)\rangle_H$ for $f\in H$. In the example of ATE, the formula is $m(w,f)=f(1,x)-f(0,x)$, so
$
\phi^{(m)}(W_i)=[\eta_j^{1/2} \{\varphi_j(1,X_i)-\varphi_j(0,X_i)\}]^{\infty}_{j=1}=\phi(1,X_i)-\phi(0,X_i).
$
The counterfactual feature map applied to observation $W_i$ replaces the observed treatment value $D_i$ with the counterfactual values $d=1$ and $d=0$.

I place two standard assumptions from kernel ridge regression analysis. The first, called the source condition, concerns an object I call $\alpha_H \in H$, which is the best RKHS approximation to the Riesz representer $\alpha_0\in L_2$.\footnote{For ATE, $\alpha_H\in\argmin_{\alpha\in H} \|\alpha-\alpha_0\|_2$ and $\alpha_0(W)=D\pi_0(X)^{-1}-(1-D)\{1-\pi_0(X)\}^{-1}$.} This assumption helps to analyze bias. The second assumption is the spectral decay condition, which quantifies the effective dimension of the features $\phi(W)$ of the RKHS $H$, and helps to analyze variance. I denote by $\mathcal{P}(b,c)$ the class of distributions that satisfy these two assumptions.

\begin{assumption}[Smoothness]\label{assumption:c}
Assume $\alpha_H\in H^c$ for some $c\in[1,2]$.
\end{assumption}

By definition of $\alpha_H \in H$ as the best kernel approximation to $\alpha_0 \in L_2$, $c\geq 1$ automatically holds in Assumption~\ref{assumption:c}. A value $c>1$ means that $\alpha_H$ is a particularly smooth element of $H$. The $L_2$ rates in this paper do not improve beyond $c=2$, reflecting the well known saturation effect of Tikhonov regularization \cite{bauer2007regularization}. 

\begin{assumption}[Spectral decay]\label{assumption:b}
(i) If $H$ is infinite dimensional, the eigenvalues $(\eta_j)$ decay at least polynomially: $ j^b \cdot \eta_j\leq \bar{B}$ for $j\geq1$, where $b\in(1,\infty)$ and $\bar{B}>0$ is a constant. If $H$ is finite dimensional, write its dimension $J$ as $J\leq\bar{B}<\infty$ and write $b=\infty$. 
(ii) If $H$ is infinite dimensional, we may further impose that the eigenvalues decay exactly polynomially: $\underline{B}  \leq j^b \cdot \eta_j \leq \bar{B}$ where $\underline{B},\bar{B}>0$ are constants.
\end{assumption}

For any bounded kernel $k$, $b\geq 1$ automatically holds in Assumption~\ref{assumption:b}(i) \cite[Lemma 10]{fischer2017sobolev}. A value $b>1$ means that the RKHS has a lower effective dimension in light of the data distribution. The $L_2$ rates in this paper improve all the way to the parameteric rate as $b\rightarrow \infty$.

The upper bound, Assumption~\ref{assumption:b}(i), applies to the  Mat\'ern, Gaussian, and other kernels with spectral decay that is polynomial, exponential, or ``better''. The lower bound, Assumption~\ref{assumption:b}(ii), applies to the Mat\'ern kernel but not the Gaussian kernel. The main results use Assumption~\ref{assumption:b}(i) only; Assumption~\ref{assumption:b}(ii) is for comparison to optimality theory.

These assumptions generalize familiar assumptions from the analysis of Sobolev spaces. For example, take $\mathcal{W}=\mathbb{R}^p$ and denote by $\mathbb{H}_2^s$ the Sobolev space with $s>p/2$ square integrable derivatives, which is an RKHS with the Mat\'ern kernel. If the RKHS used in estimation is $H=\mathbb{H}_2^s$ and the best RKHS approximation to the Riesz representer satisfies $\alpha_H\in \mathbb{H}_2^{s_0}$, then $c=s_0/s$ and  $\mathbb{H}_2^{s_0}=(\mathbb{H}_2^{s})^c$. In words, $c$ quantifies how smooth $\alpha_H$ is relative to its estimator $\tilde{\alpha}\in H$. In this RKHS, $b=2s/p$. As such, $b$ quantifies how smooth the estimator $\tilde{\alpha}\in H$ is relative to the ambient dimension, i.e. its effective dimension.

\subsection{Preview of main results}

I study a classical kernel ridge estimator $\tilde{\alpha}$ of the Riesz representer $\alpha_0$. It turns out that $\tilde{\alpha}=[\mathbb{E}_n\{\phi(W)\otimes \phi(W)^*\}+\lambda]^{-1}\mathbb{E}_n\{\phi^{(m)}(W)\}$, where I use the outer product notation $\{\phi(W)\otimes \phi(W)^*\}(\cdot)=\phi(W)\langle \phi(W),\cdot \rangle_H$ and the ridge regularization notation $\lambda=\lambda I$ with identity $I:H\rightarrow H$.  Since $\tilde{\alpha}$ involves counterfactual features $\phi^{(m)}(W)$, a notion of their effective dimension is necessary; to analyze the variance of the estimator $\tilde{\alpha}$, the counterfactual effective dimension is unavoidable.

\begin{lemma}[Main lemma]\label{lemma:preview}
    Under Assumption~\ref{assumption:cont} and weak regularity conditions (Assumptions~\ref{assumption:original} and~\ref{assumption:RKHS} below), the counterfactual effective dimension of $\phi^{(m)}(W)$ is upper bounded by $\bar{M}$ times the effective dimension of $\phi(W)$.
    See Section~\ref{sec:sketch} for details.
\end{lemma}

By Lemma~\ref{lemma:preview}, a familiar regularity condition from semiparametric theory (Assumption~\ref{assumption:cont}), together with a familiar effective dimension condition from learning theory (Assumption~\ref{assumption:b}), implies control of the counterfactual effective dimension and hence the variance of $\tilde{\alpha}$. Due to this insight, I avoid an additional approximation assumption on the counterfactual effective dimension when proving the main result; see Remark~\ref{remark:aux}.

\begin{theorem}[Main theoretical result]
    Under Assumptions~\ref{assumption:cont},~\ref{assumption:c} and~\ref{assumption:b}(i) as well as weak regularity conditions (Assumptions~\ref{assumption:original} and~\ref{assumption:RKHS} below), $\|\tilde{\alpha}-\alpha_H\|_2^2=O_p\{n^{-bc/(bc+1)}\}$ using regularization $\lambda=n^{-b/(bc+1)}$, when $b\in(1,\infty)$, $c\in(1,2]$, and $\bar{M}$ is fixed. This rate is optimal for some choices of $m$ when Assumption~\ref{assumption:b}(ii) also holds. See Section~\ref{sec:mse} for details, including results when $b=\infty$, $c=1$, or $\bar{M}_h\uparrow \infty$.
\end{theorem}

\begin{corollary}[Main corollary]
    Under weak regularity conditions and correct specification of $\hat{\gamma}$ and $\hat{\alpha}$, perhaps outside of the RKHS, if $\bar{M}$ is fixed then $\hat{\theta}$ converges to $\theta_0$ at the rate $n^{-1/2}$ and is asymptotically normal. Under mis-specification of $\hat{\gamma}$ or $\hat{\alpha}$, $\hat{\theta}$ converges to $\theta_0$, albeit at a slower rate. For pointwise approximations of nonparametric causal functions with $\bar{M}_h\uparrow\infty$, $\hat{\theta}$ converges to $\theta_0$ at the rate $(nh)^{-1/2}$ and is asymptotically normal when the bandwidth is $h=o(1)$ and $n^{-1/2}h^{-3/2}=o(1)$. See Section~\ref{sec:mse} and Appendix~\ref{sec:nonparametric}.
\end{corollary}

\begin{proposition}[A practical result]\label{prop:preview}
    The estimator $\tilde{\alpha}$ has a standalone closed form solution that can be computed from $k$ and $m$, without directly evaluating the feature maps, and that can interpolate. Kernel ridge regression and kernel ridge balancing weights are special cases, and the latter have an equivalent minimax formulation. See Section~\ref{sec:algorithm} for details.
\end{proposition}
\section{Kernel ridge Riesz representers}\label{sec:algorithm}

\subsection{General loss function}

An important departure from many works on kernel ridge balancing weights is that I allow incorrect specification by the RKHS.\footnote{See Remark~\ref{remark:spec} below.} In particular, I allow $\gamma_0 \not \in H$ and $\alpha_0 \not \in H$. For the scenario where $\alpha_0\not\in H$, I define the best kernel approximation $\alpha_H \in H$ to the Riesz representer $\alpha_0\in L_2$. I defer proofs for this section to Appendix~\ref{sec:algorithm_supp}. 

\begin{definition}[Kernel approximation to Riesz representer]\label{def:approx}
Let $\alpha_H\in\argmin_{\alpha\in H} \|\alpha-\alpha_0\|_2$.
\end{definition}

I place weak regularity conditions on the original spaces and on the RKHS $H$ that allow for further characterization of $\alpha_H$.

\begin{assumption}[Original spaces]\label{assumption:original}
$W$ takes values in a Polish space $\mathcal{W}$, i.e. a separable and completely metrizable topological space.  $Y$ takes values in $\mathcal{Y}\subset \mathbb{R}$.
\end{assumption}

\begin{assumption}[RKHS regularity]\label{assumption:RKHS}
The kernels $k$ and $k^{(m)}$ are bounded, i.e. 
    $
    \sup_{w\in\mathcal{W}}\|\phi(w)\|_H\leq \sqrt{\kappa}$ and $
    \sup_{w\in\mathcal{W}}\|\phi^{(m)}(w)\|_H\leq \sqrt{\kappa^{(m)}}$. The feature maps $\phi(w)$ and $\phi^{(m)}(w)$ are measurable. The kernel $k^{(m)}$ is characteristic \cite{sriperumbudur2010relation} in its components that vary.
\end{assumption}

The space $\mathcal{W}$ accommodates general data types, e.g. texts, images, and graphs. Commonly used kernels are bounded. Boundedness implies that the counterfactual kernel mean embedding $\mu^{(m)}=\mathbb{E}\{\phi^{(m)}(W)\}$ and the uncentered covariance operator $T=\mathbb{E}\{\phi(W)\otimes \phi(W)^*\}$ are well defined.\footnote{Here, I use outer product notation: $(a\otimes b^*)c=a \langle b,c\rangle_H$, where $b^*$ is the adjoint of $b$.} This counterfactual kernel mean embedding generalizes the well known kernel mean embedding \cite{smola2007hilbert}. It is unconditional and does not involve the outcome. 
Measurability is another standard condition. 

The characteristic property ensures that no counterfactual information is lost when approximating $\alpha_0$ as $\alpha_H$. In particular, it means that $\mathbb{P}\mapsto \mu^{(m)}$ is injective, so $\mu^{(m)}$ may be used as a sufficient statistic for $\alpha_H$, as shown in the following lemma.

\begin{lemma}[Towards a loss for KRRR]\label{lemma:loss}
    Suppose Assumptions~\ref{assumption:cont},~\ref{assumption:original}, and~\ref{assumption:RKHS} hold. Then we have $
\alpha_H\in\argmin_{\alpha\in H} \mathcal{L}(\alpha)$ where $\mathcal{L}(\alpha)=-2\langle \alpha, \mu^{(m)} \rangle_H+\langle \alpha, T\alpha\rangle_H
$.
\end{lemma}

The sufficient statistics $\mu^{(m)}\in H$ and $T\in H\otimes H^*$ are infinite dimensional generalizations of the counterfactual moment and the covariance matrix in automatic debiased machine learning (Auto-DML) \cite{chernozhukov2018global,chernozhukov2018learning}.  Auto-DML uses $p$ explicit basis functions, and its sufficient statistic are a vector in $\mathbb{R}^p$ and a matrix in $\mathbb{R}^{p \times p}$. Kernel balancing weights use the countable eigenfunctions of the kernel $k$ as infinitely many implicit basis functions.

Inspired by Lemma~\ref{lemma:loss}, I study a general loss for kernel ridge Riesz representers (KRRR) at the population level. As we will see below, its empirical analogue recovers the loss for kernel ridge balancing weights previously proposed for treatment effects. Let $\hat{\mu}^{(m)}=\mathbb{E}_n\{\phi^{(m)}(W)\}$ and $\hat{T}=\mathbb{E}_n\{\phi(W)\otimes \phi(W)^*\}$ be the empirical summary statistics. 

\begin{definition}[Loss for KRRR]\label{def:loss} Let
$
\alpha_{\lambda}=\argmin_{\alpha\in H} \mathcal{L}_{\lambda}(\alpha) $ where $ \mathcal{L}_{\lambda}(\alpha)=\mathcal{L}(\alpha)+\lambda \|\alpha\|^2_H.
$
The estimator is its empirical analogue: $
\tilde{\alpha}=\argmin_{\alpha\in H}\mathcal{L}^{n}_{\lambda}(\alpha)$ where $\mathcal{L}^{n}_{\lambda}(\alpha)=-2\langle \alpha, \hat{\mu}^{(m)} \rangle_H+\langle \alpha, \hat{T}\alpha\rangle_H+\lambda \|\alpha\|^2_H.
$
\end{definition}

The KRRR loss in Definition~\ref{def:loss} differs from Dantzig \cite{chernozhukov2018global}, Lasso \cite{chernozhukov2018learning}, and adversarial \cite{chernozhukov2020adversarial} losses for Riesz representers whose generalization errors have been previously analyzed. The KRRR loss has a simple first order condition which implies a nonasymptotic balance-variance trade-off similar to those of kernel balancing weights in the literature \cite{hazlett2020kernel,wong2018kernel,zhao2019covariate,kallus2020generalized,hirshberg2019kernel}.

\begin{lemma}[First order condition]\label{lemma:foc}
Suppose the conditions of Lemma~\ref{lemma:loss} hold. The first order conditions give
$
T\alpha_H=\mu^{(m)}$, $\alpha_{\lambda}=(T+\lambda)^{-1}\mu^{(m)}$, and $\tilde{\alpha}=(\hat{T}+\lambda)^{-1}\hat{\mu}^{(m)}.
$
\end{lemma}

\begin{corollary}[Balance-variance trade-off]\label{cor:trade}
    Suppose the conditions of Lemma~\ref{lemma:loss} hold. Then we have $\|\mathbb{E}_n\{\tilde{\alpha}(W)\phi(W)-\phi^{(m)}(W)\}\|_{H}/\|\tilde{\alpha}\|_{H}=\lambda$.
\end{corollary}

By Definition~\ref{def:loss}, a smaller regularization value $\lambda$ translates to an estimator $\tilde{\alpha}$ with a smaller bias and a larger variance. Corollary~\ref{cor:trade} illustrates that a smaller (normalized) imbalance of the features, $\|\mathbb{E}_n[\tilde{\alpha}(W)\phi(W)-\phi^{(m)}(W)]\|_{H}/\|\tilde{\alpha}\|_{H}$, is possible if and only if we pay the cost of a larger variance. In the finite sample analysis to follow, $\lambda\downarrow 0$ in a way that optimally navigates this trade-off (when rate lower bounds are known). In particular, we will not require undersmoothing for inference on the causal parameter.

\subsection{Detailed comparisons: Ridge regression and balancing weights}

Lemma~\ref{lemma:foc} has several more corollaries, which formalize the sense in which KRRR builds on and extends previous frameworks. In these corollaries, let $\tilde{\gamma}$ be the kernel ridge regression estimator. Throughout, I assume that the conditions of Lemma~\ref{lemma:loss} hold.

\begin{corollary}[Special case: Kernel ridge regression]\label{cor:krr}
 $\tilde{\alpha}=\tilde{\gamma}$ when $m(w,f)=yf(w)$.
\end{corollary}

Next, let $\tilde{\beta}\in\mathbb{R}^n$ be the kernel ridge balancing weight previously defined for average effects \cite{speckman1979minimax,kallus2020generalized,hirshberg2019augmented,hirshberg2019kernel}.

\begin{definition}[Loss for kernel ridge balancing weights]\label{def:minimax}
    Let $\tilde{\beta}=\argmin_{\beta\in\mathbb{R}^n} \tilde{\mathcal{L}}^{n}_{\lambda}(\beta)$ where $\tilde{\mathcal{L}}^{n}_{\lambda}(\beta)=[\sup_{f\in H:\|f\|_H\leq 1}\frac{1}{n}\sum_{i=1}^n \{\beta_i f(W_i)-m(W_i,f)\}]^2+n^{-1}\lambda \beta^\top\beta.
$
\end{definition}

\begin{corollary}[Special case: Kernel ridge balancing weights; eq. 8 of \cite{bruns2023augmented}]\label{cor:hirshberg}
    On the training data, $\tilde{\alpha}(W_i)=\tilde{\beta}_i$. Appendix D.3 of \cite{hirshberg2019kernel} studies $\frac{1}{n}\sum_{i=1}^n \{\tilde{\beta}_i-\alpha_0(D_i,X_i)\}^2$, i.e. training error.
\end{corollary}

In summary, KRRR nests kernel ridge regression and kernel ridge balancing weights, unlike the adversarial Riesz representer of \cite{chernozhukov2020adversarial}. The connection to kernel ridge regression foreshadows the new generalization error guarantees for KRRR. 

Next, I recap known equivalences for the estimation of treatment effects \cite{speckman1979minimax,kallus2020generalized,hirshberg2019kernel} and more generally for the estimation of causal parameters within the class of Assumption~\ref{assumption:cont} \cite{bruns2023augmented}. With $\hat{\alpha}$ taken to be KRRR $\tilde{\alpha}$, these equivalences only hold when the regression estimator $\hat{\gamma}$ is kernel ridge regression $\tilde{\gamma}$, and when the same observations are used for $\tilde{\gamma}$, $\tilde{\alpha}$, and the causal parameter.

\begin{corollary}[Equivalence in a special case; Proposition 27 of \cite{kallus2020generalized}]\label{cor:equal}
When $\tilde{\gamma}$ is kernel ridge regression and observations are reused for $\tilde{\gamma}$, $\tilde{\alpha}$, and the causal parameter, $\mathbb{E}_n\{m(W,\tilde{\gamma})\}=\mathbb{E}_n\{Y\tilde{\alpha}(W)\}$.
\end{corollary}

\begin{corollary}[Non-equivalence in general]\label{cor:non-equal}
    For a generic regression estimator $\hat{\gamma}\neq\tilde{\gamma}$, we have $\mathbb{E}_n\{m(W,\hat{\gamma})\}\neq \mathbb{E}_n\{Y\tilde{\alpha}(W)\}$. For any regression estimator $\hat{\gamma}$ (including $\tilde{\gamma}$), if $\hat{\gamma}$ or $\tilde{\alpha}$ is estimated on a held out sample, then the $\mathbb{E}_n\{m(W,\hat{\gamma})\}\neq \mathbb{E}_n\{Y\tilde{\alpha}(W)\}$.
\end{corollary}

When $\gamma_0\not \in H$, alternative regression estimators lead to better guarantees for semiparametric inference; see Section~\ref{sec:semi} below.

\subsection{Standalone closed form solution}

In summary, the standalone closed form for KRRR is not necessary for causal estimation in the special case of Corollary~\ref{cor:equal}. However, Corollary~\ref{cor:non-equal} shows that if $\hat{\gamma}\neq \tilde{\gamma}$, or if $\hat{\gamma}$ or $\tilde{\alpha}$ is estimated on a held out sample, then a standalone closed form for KRRR that can be evaluated away from the training data is necessary for causal estimation. I derive it below using techniques of \cite{chernozhukov2020adversarial}, as a secondary contribution.

Even for the special case of Corollary~\ref{cor:equal}, this paper's primary contribution is the population $L_2$ rate $\|\tilde{\alpha}-\alpha_H\|_2$ for kernel ridge balancing weights (and more generally KRRR) in Section~\ref{sec:mse}. Generalization error is essential for appealing to strong semiparametric guarantees that allow some mis-specification and justify inference on causal functions.

I use some additional notation for features. Define the operator $\Phi:H\rightarrow \mathbb{R}^n$ with $i$th component $\langle \phi(W_i),\cdot \rangle_H$ and likewise the operator $\Phi^{(m)}:H\rightarrow \mathbb{R}^n$ with $i$th component $\langle \phi^{(m)}(W_i),\cdot \rangle_H$. Concatenate $\Phi$ and $\Phi^{(m)}$ as $\Psi:H\rightarrow \mathbb{R}^{2n}$, with adjoint $\Psi^*:\mathbb{R}^{2n}\rightarrow H$.

\begin{lemma}[Standalone closed form exists; c.f. Lemma 2 of \cite{chernozhukov2020adversarial}]\label{lemma:represent}
    Suppose the conditions of Lemma~\ref{lemma:loss} hold. There exists some $\rho\in\mathbb{R}^{2n}$ such that $\tilde{\alpha}=\Psi^*\rho$.
\end{lemma}

Whereas kernel ridge regression has a solution in $\mathbb{R}^n$ \cite{kimeldorf1971some,scholkopf2001generalized}, KRRR has a solution in $\mathbb{R}^{2n}$ because of the counterfactual features. Intuitively, causal inference involves reasoning about $n$ actual observations and $n$ counterfactual observations.\footnote{The conjecture in \cite[Appendix D.2]{hirshberg2019augmented} that kernel ridge balancing weights may have a representation $\rho\in \mathbb{R}^n$, such that $\tilde{\alpha}=\Phi^*\rho$, does not appear to be supported in general.} 

The familiar kernel matrix $K^{(1)}=\Phi\Phi^*\in\mathbb{R}^{n\times n}$ encodes inner products between actual observations. The extended kernel matrix $K=\Psi \Psi^* \in \mathbb{R}^{2n\times 2n}$ encodes inner products between actual observations and counterfactual observations. Each entry of $K$ is computed from the kernel $k$ and formula $m$ as discussed in Section~\ref{sec:problem_rkhs}. For intuition, write
$$
\Psi=\begin{bmatrix} \Phi\\ \Phi^{(m)}\end{bmatrix},\quad K=\begin{bmatrix} K^{(1)} & K^{(2)} \\ K^{(3)}  & K^{(4)} \end{bmatrix}=\begin{bmatrix} \Phi \Phi^* & \Phi \{\Phi^{(m)}\}^* \\ \Phi^{(m)}\Phi^* & \Phi^{(m)}  \{\Phi^{(m)}\}^*  \end{bmatrix}.
$$

\begin{proposition}[Standalone closed form for KRRR]\label{prop:closed}
    Suppose the conditions of Lemma~\ref{lemma:loss} hold.  Given observations $(W_i)$ and evaluation location $w$, as well as the formula $m$, kernel $k$, and regularization $\lambda$, the KRRR closed form is as follows.
\begin{enumerate}
    \item Calculate $K^{(j)}\in\mathbb{R}^{n\times n}$ as defined above, for $j\in\{1,2,3,4\}$.
    \item Calculate $\Omega^{2n\times 2n}$ and $u(w),v\in\mathbb{R}^{2n}$ by
    $$
    \Omega=\begin{bmatrix} K^{(1)}K^{(1)} & K^{(1)}K^{(2)}\\ K^{(3)}K^{(1)} & K^{(3)}K^{(2)} \end{bmatrix},\; v=\begin{bmatrix} K^{(2)} \\  K^{(4)} \end{bmatrix} \mathbbm{1}_{n},\; u_i(w)=\begin{cases} k(W_i,w) & \text{ if }i\in \{1,...,n\}\\ k_m(W_i,w)& \text{ if }i\in \{n+1,...,2n\}\end{cases}
    $$
     where $\mathbbm{1}_{n}\in\mathbb{R}^{n}$ is a vector of ones, $k(w,w')=\langle \phi(w),\phi(w')\rangle_H$, and $k_m(w,w')=\langle \phi^{(m)}(w),\phi(w')\rangle_H$.
    \item Set
    $
    \tilde{\alpha}(w)=v^{\top}(\Omega+n\lambda K)^{-1} u(w).
    $
\end{enumerate}
\end{proposition}

The standalone closed form solution for kernel ridge balancing weights, and KRRR more generally, appears to have been previously unknown. Proposition~\ref{prop:closed} demonstrates that it can be easily computed using only the formula $m$ and kernel $k$, without directly evaluating the feature maps $\phi(w)$ or $\phi^{(m)}(w)$. In this sense, it is a practical contribution for inference. See Appendix~\ref{sec:details} for its specialization to Example~\ref{ex:het}.
\section{Generalization error with mis-specification}\label{sec:mse}

\subsection{Main result}

I now provide the main result: population $L_2$ rates for KRRR that coincide with the optimal rate, where optimal rates are known. I study the class of Riesz representers defined by Assumption~\ref{assumption:cont}, imposing familiar smoothness and spectral decay conditions for the RKHS setting defined by Assumptions~\ref{assumption:c} and~\ref{assumption:b}. I also impose mild regularity conditions that help in the algorithm derivation, defined by Assumptions~\ref{assumption:original} and~\ref{assumption:RKHS}. 

For clarity, I write the population $L_2$ rate for KRRR, estimated from the observations $(W_i)_{i\in[n]}$, in terms of $\mathcal{R}(\tilde{\alpha})=\mathbb{E}[\{\tilde{\alpha}(W)-\alpha_H(W)\}^2|(W_i)_{i\in[n]}]$. It is the generalization error for a test observation $W$, allowing for mis-specification.

\begin{theorem}[Upper bound for KRRR]\label{theorem:riesz}
    If Assumptions~\ref{assumption:cont},~\ref{assumption:c},~\ref{assumption:b}(i),~\ref{assumption:original}, and~\ref{assumption:RKHS} hold with $\bar{M}$ bounded above, 
then
$
\lim_{\tau\rightarrow \infty} \lim\sup_{n\rightarrow\infty} \sup_{p\in\mathcal{P}(b,c)} \mathbb{P}_{(W_i)\sim p^{n}}\{\mathcal{R}(\tilde{\alpha})>\tau \cdot  r_n\}=0
$, 
where
$$
r_n=\begin{cases} n^{-1} & \text{ if }\quad b=\infty \text{ and } \lambda=n^{-\frac{1}{2}}; \\
n^{-\frac{bc}{bc+1}} & \text{ if }\quad b\in(1,\infty),\; c\in(1,2]\text{ and } \lambda=n^{-\frac{b}{bc+1}}; \\
\ln^{\frac{b}{b+1}}(n)\cdot n^{-\frac{b}{b+1}} & \text{ if }\quad b\in(1,\infty),\; c=1\text{ and } \lambda=\ln^{\frac{b}{b+1}}(n)\cdot n^{-\frac{b}{b+1}}.
\end{cases}
$$
\end{theorem}

\begin{remark}[Causal functions]
    When the causal parameter has a formula of the form $m_h(w,f)=\ell_h(w)\check{m}(w,f)$, as in Example~\ref{ex:het}, $\bar{M}_h\uparrow \infty$. In such case, I use the estimator $\hat{\alpha}_h(W)=\ell_h(W)\tilde{\alpha}(W)$, where $\tilde{\alpha}$ is KRRR for the formula $\check{m}(w,f)$. When $\tilde{\alpha}$ has the rate $r_n$, $\hat{\alpha}_h$ has the rate $h^{-2}r_n$ under weak regularity conditions. See Appendix~\ref{sec:nonparametric} for details.
\end{remark}

I defer the proof to Section~\ref{sec:sketch}. I defer the proofs of corollaries to Appendix~\ref{sec:mse_details}. 

Theorem~\ref{theorem:riesz} does not require Assumption~\ref{assumption:b}(ii). Moreover, as an intermediate step, I prove a nonasymptotic bound (Proposition~\ref{prop:abstract}) without imposing Assumptions~\ref{assumption:c} and~\ref{assumption:b}(i).

When the RKHS is finite dimensional, i.e. $b=\infty$ in Assumption~\ref{assumption:b}, KRRR achieves the parametric rate $r_n=n^{-1}$. When the RKHS is infinite dimensional with at least polynomial spectral decay, i.e. $b<\infty$ in Assumption~\ref{assumption:b}, KRRR achieves a familiar rate $r_n=n^{-\frac{bc}{bc+1}}$. This rate depends on $c$ in Assumption~\ref{assumption:c}, which quantifies the smoothness of the best kernel approximation $\alpha_H \in H$ to the Riesz representer $\alpha_0 \in L_2$. If no additional smoothness of $\alpha_H$ is known, i.e. all we have is that $\alpha_H\in H$ by construction, then $c=1$ and the rate has an extra logarithmic factor.

The rate $r_n=n^{-\frac{bc}{bc+1}}$ generalizes the familiar rate from Sobolev analysis. For example, take $\mathcal{W}=\mathbb{R}^p$ and denote by $\mathbb{H}_2^s$ the Sobolev space with $s>p/2$ square integrable derivatives, as before. Suppose that KRRR $\tilde{\alpha}$ is estimated using the RKHS $H=\mathbb{H}_2^s$. Suppose that $\alpha_H\in \mathbb{H}_2^{s_0}$, i.e. the best kernel approximation to the Riesz representer $\alpha_0\in L_2$ has $s_0>s$ square integrable derivatives. The population $L_2$ rate for $\tilde{\alpha}$ is then $r_n=n^{-\frac{2s_0}{2s_0+p}}$. When $s_0=s$, which is a lower bound on $s_0$ by construction, $r_n=\ln(n)^{\frac{2s}{2s+p}}\cdot n^{-\frac{2s}{2s+p}}$. Theorem~\ref{theorem:riesz} applies to settings beyond Sobolev spaces over Euclidean domains.

\begin{corollary}[Special case: Kernel ridge regression; Theorem 1 of \cite{caponnetto2007optimal}]\label{cor:cap}
    Suppose the conditions of Theorem~\ref{theorem:riesz} hold, replacing Assumption~\ref{assumption:cont} with $|Y|\leq \bar{Y}$ almost surely. Then
$
\lim_{\tau\rightarrow \infty} \lim\sup_{n\rightarrow\infty} \sup_{p\in\mathcal{P}(b,c)} \mathbb{P}_{(W_i)\sim p^{n}}\{\mathcal{R}(\tilde{\gamma})>\tau \cdot  r_n\}=0.
$
\end{corollary}

\begin{remark}[Excess risk]
    Some works on kernel ridge regression state results in terms of the excess risk $\mathcal{E}(\tilde{\gamma})-\inf_{\gamma\in H}\mathcal{E}(\gamma)$, where $\mathcal{E}(f)=\mathbb{E}[\{Y-f(W)\}^2]$. This is equivalent to  $\mathcal{R}(\tilde{\gamma})=\mathbb{E}[\{\tilde{\gamma}(W)-\gamma_H(W)\}^2|(W_i)_{i\in[n]}]$ when $\gamma_H\in \argmin_{\gamma \in H} \mathcal{E}(\gamma)$ exists, which is assumed in those works and in this one. See e.g. \cite[eq. 24]{caponnetto2007optimal} and \cite[Lemma 12]{fischer2017sobolev}.
\end{remark}

Just as KRRR generalizes kernel ridge regression, Theorem~\ref{theorem:riesz} generalizes \cite[Theorem 1]{caponnetto2007optimal}. The contribution is meaningful because KRRR also generalizes the classical kernel ridge balancing weights, for which generalization error was not fully articulated. In settings where lower bounds are known, the upper bounds in Theorem~\ref{theorem:riesz} often achieve them.

\begin{lemma}[Lower bound; Theorem 2 of \cite{caponnetto2007optimal}]\label{lemma:lower}
    Suppose the conditions of Corollary~\ref{cor:cap} hold as well as Assumption~\ref{assumption:b}(ii), $b\in (1,\infty)$, and $c\in [1,2]$. Then 
    $
    \lim_{\tau\rightarrow 0} \lim\inf_{n\rightarrow\infty} \inf_{\hat{\gamma}} \sup_{p\in \mathcal{P}(b,c)}\mathbb{P}_{(W_i)\sim p^{n}}\{\mathcal{R}(\hat{\gamma})>\tau \cdot  n^{-\frac{bc}{bc+1}}\}=1.
    $
\end{lemma}

\begin{corollary}[Optimality]\label{cor:optimal}
When $b\in(1,\infty)$ and $c\in(1,2]$, Theorem~\ref{theorem:riesz} matches Lemma~\ref{lemma:lower}. When $b\in(1,\infty)$ and $c=1$, they match up to a log factor.
\end{corollary}

\subsection{Semiparametric inference beyond the RKHS}\label{sec:semi}

To showcase the role of Theorem~\ref{theorem:riesz} in semiparametric inference, I quote a result from the targeted and debiased machine learning literature. Unlike many previous works on kernel ridge balancing weights, inference allows $\gamma_0\not\in H$. It also allows for sample splitting, which is helpful when the regression estimator $\hat{\gamma}$ is not kernel ridge regression, and is instead estimated in a function space with high entropy. Theorem~\ref{theorem:riesz} is directly compatible with the quoted result, unlike previous guarantees for kernel ridge balancing weights. In this sense, KRRR provides debiased kernel methods. 

To state the result, denote the doubly robust estimator with sample splitting as $\hat{\theta}$, with the well known analytic variance estimator $\hat{\sigma}^2$.

\begin{definition}[Debiased machine learning]\label{def:target}
Given a sample $(Y_i,W_i)_{i\in[n]}$, partition the sample into folds $(I_{\ell})_{\ell\in [L]}$. Denote by $I_{\ell}^c$ the complement of $I_{\ell}$.
\begin{enumerate}
    \item For each fold $\ell$, estimate $\hat{\gamma}_{\ell}$ and $\hat{\alpha}_{\ell}$ from observations in $I_{\ell}^c$.
    \item Estimate $\theta_0$ as
    $
  \hat{\theta}=n^{-1}\sum_{\ell=1}^L\sum_{i\in I_{\ell}} [m(W_i,\hat{\gamma}_{\ell})+\hat{\alpha}_{\ell}(W_i)\{Y_i-\hat{\gamma}_{\ell}(W_i)\}]
    $.
    \item Estimate $\hat{\sigma}^2=n^{-1}\sum_{\ell=1}^L\sum_{i\in I_{\ell}} [m(W_i,\hat{\gamma}_{\ell})+\hat{\alpha}_{\ell}(W_i)\{Y_i-\hat{\gamma}_{\ell}(W_i)\}-\hat{\theta}]^2
    $.
    \item Construct the $(1-a) 100$\% confidence interval as
    $
    \hat{\theta}\pm c_{a}\hat{\sigma} n^{-1/2}$, where $c_{a}$ is the $1-a/2$ quantile of the standard Gaussian.
\end{enumerate}
\end{definition}

Let $\gamma_{\mathcal{G}}$ be the best approximation to $\gamma_0$ in some space $\mathcal{G}$, which may not be an RKHS. Similarly, let $\alpha_{\mathcal{A}}$ be the best approximation to $\alpha_0$ in some space $\mathcal{A}$. Slightly abusing notation, I now let $\mathcal{R}(\hat{\gamma})=\mathbb{E}[\{\hat{\gamma}(W)-\gamma_{\mathcal{G}}(W)\}^2|(W_i)_{i\in[n]}]$ and $\mathcal{R}(\hat{\alpha})=\mathbb{E}[\{\hat{\alpha}(W)-\alpha_{\mathcal{A}}(W)\}^2|(W_i)_{i\in[n]}]$.\footnote{In Theorem~\ref{theorem:riesz} above, $\mathcal{A}=H$. In Corollary~\ref{cor:cap} above, $\mathcal{G}=H$.}

To quote the result, I introduce some additional notation. Let $\psi(W,\theta,\gamma,\alpha)=m(W,\gamma)+\alpha(W)\{Y-\gamma(W)\}-\theta$, so that $\psi_0(W)=\psi(W,\theta_0,\gamma_0,\alpha_0)$ is the influence function with moments  $\sigma^2=\mathbb{E}\{\psi_0(W)^2\}$, $\zeta^3=\mathbb{E}\{\psi_0(W)^3\}$, and $\chi^4=\mathbb{E}\{\psi_0(W)^4\}$.

\begin{lemma}[Inference; Corollary 1 of \cite{chernozhukov2021simple}]\label{lemma:dml}
Suppose Assumption~\ref{assumption:cont} holds. Suppose the following regularity conditions:
$
\mathbb{E}[\{Y-\gamma_0(W)\}^2 \mid W]\leq \bar{\sigma}^2$, $ \|\alpha_0\|_{\infty}\leq\bar{\alpha}$, $  \|\hat{\alpha}\|_{\infty}\leq\bar{\alpha}'$, and $\left\{\left(\zeta/\sigma\right)^3+\chi^2\right\}n^{-1/2}\rightarrow0.
$
Suppose the following learning rate conditions: 
\begin{enumerate}
    \item $\left(\bar{M}^{1/2}+\bar{\alpha}/\sigma+\bar{\alpha}'\right)\{\mathcal{R}(\hat{\gamma})+\|\gamma_{\mathcal{G}}-\gamma_0\|_2^2\}^{1/2}=o_p(1)$;
     $\bar{\sigma}\{\mathcal{R}(\hat{\alpha})+\|\alpha_{\mathcal{A}}-\alpha_0\|_2^2\}^{1/2}=o_p(1)$;
    \item $[n \{\mathcal{R}(\hat{\gamma})+\|\gamma_{\mathcal{G}}-\gamma_0\|_2^2\} \{\mathcal{R}(\hat{\alpha})+\|\alpha_{\mathcal{A}}-\alpha_0\|_2^2\}]^{1/2}/\sigma =o_p(1)$. 
\end{enumerate} 
Then  
$\sigma^{-1}n^{1/2}(\hat{\theta}-\theta_0)\leadsto\mathcal{N}(0,1)$ and $ \mathbb{P}\left\{\theta_0 \in \left(\hat{\theta}\pm c_a\hat{\sigma} n^{-1/2} \right)\right\}\rightarrow 1-a
$.
\end{lemma}

Using weak regularity conditions, population $L_2$ rates $\mathcal{R}(\hat{\gamma})$ and $\mathcal{R}(\hat{\alpha})$, and approximation errors $\|\gamma_{\mathcal{G}}-\gamma_0\|_2$ and $\|\alpha_{\mathcal{A}}-\alpha_0\|_2$, the estimator $\hat{\theta}$ is consistent and asymptotically Gaussian at the rate $n^{-1/2}\sigma$. Its confidence interval includes $\theta_0$ with probability approaching the nominal level. Importantly, $\hat{\gamma}$ does not have to be kernel ridge regression. 

\begin{remark}[Mis-specification of features]\label{remark:spec}
    Crucially, $\gamma_0$ does not need to be well specified in the RKHS $H$. Instead, $\|\gamma_{\mathcal{G}}-\gamma_0\|_2$ must vanish for some $\mathcal{G}$ that may not be $H$. Previous works on classical kernel ridge balancing weights, which provide rates similar to $n^{-1/2}\sigma$ for the causal parameter, typically require $\gamma_0\in H$; see e.g. \cite[Assumption 2]{hazlett2020kernel}, \cite[Assumption 3]{wong2018kernel}, \cite[Assumption 2]{zhao2019covariate}, \cite[Theorem 18(a)]{kallus2020generalized}, \cite[Assumption 3]{nie2021quasi}, \cite[Assumption 3]{hirshberg2019kernel}, \cite[Assumption 4]{wang2021estimation} and \cite[Theorem 2]{mou2023kernel}. 
    
    \cite[eq. 10]{hirshberg2019augmented} requires $\|\hat{\gamma}-\gamma_0\|_H=O_p(1)$ and $\mathbb{E}_n[\{\hat{\gamma}(W)-\gamma_0(W)\}^2]=o_p(1)$, which is satisfied by correctly specified kernel ridge regression, i.e. $\hat{\gamma}=\tilde{\gamma}$ and $\gamma_0\in H$ \cite[Remark 2]{hirshberg2019augmented}.
    
    To use Lemma~\ref{lemma:dml} with classical kernel ridge balancing weights, allowing $\gamma_0\not \in H$, we require a rate $\mathcal{R}(\hat{\alpha})$, which is the main result of this paper (Theorem~\ref{theorem:riesz}).
\end{remark}

As written, Lemma~\ref{lemma:dml} applies to pointwise approximations of nonparametric causal functions such as heterogeneous treatment effects (Example~\ref{ex:het}). In such case, $\sigma=\sigma_h \asymp  h^{-1/2}$, so the rate of Gaussian approximation is $(nh)^{-1/2}$. In practice, I take $h\asymp n^{-1/5}$ so the rate becomes $n^{-2/5}$; see Section~\ref{sec:experiments}.
This nonparametric rate is slower than the $n^{-1/2}$ obtained for regular cases such as ATE, similar to results of e.g. \cite{van2018cv,nie2021quasi,foster2023orthogonal,kennedy2023towards} and many references therein, which could have been quoted instead. The point of Lemma~\ref{lemma:dml} is not to provide new semiparametric theory but to demonstrate how the main result of this paper (Theorem~\ref{theorem:riesz}) makes general semiparametric guarantees immediate. See Appendix~\ref{sec:nonparametric} for more details on nonparametric causal functions.\footnote{When appealing to Theorem~\ref{theorem:riesz} in the regular case, the rate for $\mathcal{R}(\hat{\alpha})$ is $r_n$. When appealing to Theorem~\ref{theorem:riesz} for causal functions, the rate for $\mathcal{R}(\hat{\alpha}_h)$ is $h^{-2}r_n$.}

\begin{remark}[Rate-optimal ridge regularization]
Lemma~\ref{lemma:dml} does not require undersmoothing of the regularization in $\hat{\gamma}$ or $\hat{\alpha}$ in order to prove inference for $\hat{\theta}$, departing from e.g. \cite{hirshberg2019kernel,mou2023kernel}. Instead, it requires $L_2$ rates for $\hat{\gamma}$ and $\hat{\alpha}$, the latter of which Theorem~\ref{theorem:riesz} provides. These rates are obtained with rate-optimal ridge regularization when optimality is known, as formalized in Corollary~\ref{cor:optimal}. When $b<\infty $ and $c=1$, optimal ridge regularization (up to log factors) is $\lambda=n^{-\frac{b}{b+1}}$.

For comparison, when $b<\infty$, \cite[Assumption 3]{hirshberg2019kernel} imposes $\lambda \ll n^{-1}$, which undersmooths. 
The regularization in \cite[Theorem 1]{hirshberg2019augmented} is $\lambda\asymp n^{-1}$, and the authors ``generally recommend'' $\lambda=\bar{\sigma}^2 n^{-1}$, which undersmooths. The discussion following \cite[Theorem 2]{hirshberg2019augmented} considers alternative choices, e.g. $\lambda\asymp n^{-\frac{b}{2}}$ when $b<\infty$ \cite[Lemma 6]{hirshberg2019kernel} which seems to undersmooth for $b>1$.\footnote{Future work may characterize whether rate-optimal ridge regularization of KRRR is compatible with \cite[eq. 19]{hirshberg2019augmented}. \cite[Appendix B.2]{hirshberg2019augmented} demonstrates rate-optimal regularization for H\"older spaces rather than RKHSs.}
See \cite[Section 7]{bruns2023augmented} for a comprehensive and insightful discussion when $\hat{\gamma}=\tilde{\gamma}$ and $\hat{\alpha}=\tilde{\alpha}$.
\end{remark}

The familiar doubly robust guarantee also holds, where either $\hat{\gamma}$ or $\hat{\alpha}$ may be mis-specified yet $\hat{\theta}$ remains consistent, albeit at a slower rate than $n^{-1/2}\sigma$. In particular, either $\|\gamma_{\mathcal{G}}-\gamma_0\|_2$ or $\|\alpha_{\mathcal{A}}-\alpha_0\|_2$ may be non-vanishing. Denote the mis-specified second moment as $\sigma_{\textsc{mis}}^2=\mathbb{E}\{\psi(W,\theta_0,\gamma_{\mathcal{G}},\alpha_{\mathcal{A}})^2\}$.

\begin{proposition}[Consistency under mis-specification]\label{prop:mis}
Suppose Assumption~\ref{assumption:cont} holds. Suppose the following regularity conditions:
$
\mathbb{E}[\{Y-\gamma_{\mathcal{G}}(W)\}^2 \mid W]\leq \bar{\sigma}^2$, $  \|\hat{\alpha}\|_{\infty}\leq\bar{\alpha}'$, and $\sigma_{\textsc{mis}} n^{-1/2}\rightarrow0
$. 
Suppose the following learning rate conditions: 
 $\left(\bar{M}^{1/2}+\bar{\alpha}'\right)\mathcal{R}(\hat{\gamma})^{1/2}=o_p(1)$ and
   $\bar{\sigma}\mathcal{R}(\hat{\alpha})^{1/2}=o_p(1)$. 
If either $\gamma_{\mathcal{G}}=\gamma_0$ or $\alpha_{\mathcal{A}}=\alpha_0$, then
$\hat{\theta}=\theta_0+o_p(1)$.
\end{proposition}

See Appendix~\ref{sec:mse_details} for the proof and the exact nonasymptotic rate of convergence, which depends on $\mathcal{R}(\hat{\gamma})$ and $\mathcal{R}(\hat{\alpha})$. Previous works on kernel ridge balancing weights typically require $\gamma_0\in H$. By contrast, Proposition~\ref{prop:mis} tolerates nonvanishing $\|\gamma_{\mathcal{G}}-\gamma_0\|_2$ or  nonvanishing $\|\alpha_{\mathcal{A}}-\alpha_0\|_2$, where neither $\mathcal{G}$ nor $\mathcal{A}$ may be $H$. To achieve this robustness to mis-specification for KRRR, we require a rate $\mathcal{R}(\hat{\alpha})$, i.e. the main result of this paper.

Again, as written, Proposition~\ref{prop:mis} applies to pointwise approximations of nonparametric causal functions. See Appendix~\ref{sec:nonparametric} for details.

Proposition~\ref{prop:mis} summarizes the nonasymptotic Proposition~\ref{prop:non}, which is a modest refinement of the celebrated double robustness guarantee. It is merely for illustrative purposes; stronger asymptotic results are available in the literature. For inference under mis-specification, see e.g. \cite{van2014targeted,benkeser2017doubly,dukes2021doubly}.
\section{Proof via the counterfactual effective dimension}\label{sec:sketch}

I now prove the main result, drawing on integral operator techniques for kernel ridge regression from \cite{smale2007learning,caponnetto2007optimal,fischer2017sobolev} and many important references therein. The main innovation is the counterfactual effective dimension.

The steps are as follows: (i) translate the desired statement from generalization error to $H$ norm (Lemma~\ref{lemma:norms}); (ii) characterize high probability events via concentration in $H$ (Proposition~\ref{prop:high}); (iii) use these high probability events to prove a nonasymptotic bound in terms of standard learning theory quantities (Proposition~\ref{prop:abstract}). 

Among the high probability events is one that involves the counterfactual features $\phi^{(m)}(W)$. It pertains to the variance of the KRRR estimator $\tilde{\alpha}$. To justify this high probability event, I bound the counterfactual effective dimension of $\phi^{(m)}(W)$ in terms of the actual effective dimension of $\phi(W)$ (Lemma~\ref{lemma:prior_new}).

The main result (Theorem~\ref{theorem:riesz}) then follows from simplifying the nonasymptotic bound (Proposition~\ref{prop:abstract}) using known bounds on the learning theory quantities that hold under smoothness (Assumption~\ref{assumption:c}) and spectral decay (Assumption~\ref{assumption:b}) conditions. The latter controls the actual effective dimension. I defer proofs of lemmas to Appendix~\ref{sec:sketch_details}.

\begin{lemma}[From generalization error to $H$ norm; c.f. Lemma 12 of \cite{fischer2017sobolev}]\label{lemma:norms} Suppose the conditions of Lemma~\ref{lemma:loss} hold. Then
$
\mathcal{R}(\tilde{\alpha})=\|T^{\frac{1}{2}}(\tilde{\alpha}-\alpha_H)\|^2_H.
$
\end{lemma}

\begin{definition}[Standard learning theory quantities]\label{def:quantities}
Define the residual $\mathcal{A}(\lambda)=\|T^{1/2}(\alpha_{\lambda}-\alpha_H)\|_H^2$, the reconstruction error $ \mathcal{B}(\lambda)= \|\alpha_{\lambda}-\alpha_H\|^2_H$, and the actual effective dimension $\mathcal{N}(\lambda)= Tr\{(T+\lambda)^{-1}T\} $.
\end{definition}

\begin{definition}[New learning theory quantity]\label{def:quantity_new}
   Define the counterfactual effective dimension $\mathcal{N}^{(m)}(\lambda)=Tr\{(T+\lambda)^{-1}T^{(m)}\}$, where $T^{(m)}=\mathbb{E}[\phi^{(m)}(W)\otimes \{\phi^{(m)}(W)\}^*]$ is the counterfactual covariance operator.
\end{definition}

\begin{lemma}[New learning theory bound]\label{lemma:prior_new}
Suppose the conditions of Lemma~\ref{lemma:loss} hold.  Then 
$
\mathcal{N}^{(m)}(\lambda)\leq \bar{M} \cdot \mathcal{N}(\lambda).
$
\end{lemma}

\begin{remark}[Avoiding an auxiliary approximation assumption]\label{remark:aux}
Lemma~\ref{lemma:prior_new} demonstrates, by a simple argument, that a standard assumption in semiparametric theory (Assumption~\ref{assumption:cont}) implies that the counterfactual effective dimension (Definition~\ref{def:quantity_new}) is not much greater than the actual effective dimension (Definition~\ref{def:quantities}). This connection between semiparametric theory and learning theory appears to be new, and it powers the main result.

Previous works that study generalization error, for different estimators of Riesz representers, appear to place auxiliary approximation assumptions to limit the counterfactual effective dimension, e.g. \cite[Assumption SC(a)]{chernozhukov2018global}, \cite[Assumption 2]{chernozhukov2018learning}, and \cite[Assumption 2]{chernozhukov2020adversarial}.
\end{remark}

\begin{definition}[High probability events in $H$]
Let $\|\cdot\|_{\mathcal{L}_2(H,H)}$ be the Hilbert-Schmidt norm for operators that map from $H$ to $H$. Define the events
{ \small 
\begin{align*}
    \mathcal{E}_1&=\left[\|(T+\lambda)^{-1}(\hat{T}-T)\|_{\mathcal{L}_2(H,H)}\leq 2\ln(6/\delta)\left\{\frac{2\kappa}{\lambda n}+\sqrt{\frac{\kappa\mathcal{N}(\lambda)}{\lambda n}}\right\}\right], \\
    \mathcal{E}_2&=\left[\|(T-\hat{T})(\alpha_{\lambda}-\alpha_H)\|_H\leq 
    2\ln(6/\delta)\left\{\frac{2\kappa\sqrt{\mathcal{B}(\lambda)}}{n}+\sqrt{\frac{\kappa\mathcal{A}(\lambda)}{n}}\right\}
    \right],  \\
    \mathcal{E}_3&=\left[\|(T+\lambda)^{-\frac{1}{2}}(\hat{\mu}^{(m)}-\hat{T}\alpha_H)\|_H
    \leq
    2\ln(6/\delta)\left\{\frac{1}{n}\sqrt{\Upsilon^2\frac{\kappa'}{\lambda}}+\sqrt{\frac{\Sigma^2\mathcal{N}(\lambda)}{n}}\right\}
    \right],
\end{align*}
}
where
$
   \Upsilon=2(1+\sqrt{\kappa}\|\alpha_H\|_H)$, $\Sigma^2=2(\bar{M}+\kappa \|\alpha_H\|^2_H)$, and $\kappa'=\max\{\kappa,\kappa^{(m)}\}.
    $
\end{definition}

\begin{proposition}[High probability events in $H$]\label{prop:high}
Suppose the conditions of Lemma~\ref{lemma:loss} hold. Then
$
\mathbb{P}(\mathcal{E}_j^c)\leq \delta/3$ for each $j\in\{1,2,3\}.
$
\end{proposition}

The proof of Proposition~\ref{prop:high} appeals to Lemma~\ref{lemma:prior_new} in order to handle $\mathcal{E}_3$, which contains the counterfactual features via the counterfactual mean embedding $\hat{\mu}^{(m)}=\mathbb{E}_n\{\phi^{(m)}(W)\}$.

\begin{proposition}[Nonasymptotic bound]\label{prop:abstract}
Suppose the conditions of Lemma~\ref{lemma:loss} hold. If $n$ is sufficiently large that
$n\geq 192 \ln^2(6/\delta) \kappa' \mathcal{N}(\lambda) \lambda^{-1}$ and $\lambda\leq \|T\|_{op},
$
then with probability $1-\delta$,
$
\|T^{1/2}(\tilde{\alpha}-\alpha_H)\|^2_H\leq 96 \ln^2(6/\delta) \left\{\mathcal{A}(\lambda)+\frac{\kappa^2\mathcal{B}(\lambda)}{n^2\lambda}+\frac{\kappa \mathcal{A}(\lambda)}{ n\lambda}+\frac{\kappa' \Upsilon^2}{n^2\lambda}+\frac{\Sigma^2 \mathcal{N}(\lambda)}{n}\right\}.
$
\end{proposition}

 The proof of Proposition~\ref{prop:abstract} appeals to Proposition~\ref{prop:high} and the union bound.

\section{Simulated and real data analysis}\label{sec:experiments}

\subsection{Nominal pointwise coverage of heterogeneous effects}

I present coverage simulations for heterogeneous treatment effects (Example~\ref{ex:het}) viewed as a causal function. To lighten notation, I let $V=X_1$ and I write the heterogeneous treatment effects as $\textsc{cate}(v)$.

I follow the simulation design of \cite{abrevaya2015estimating}, where $\textsc{cate}(v)=v(1+2v)^2(v-1)^2$ is the function on which we aim to conduct inference using observations of the outcome $Y$, binary treatment $D$, and covariates $X$. The covariate of interest $V\in [-0.5,0.5]$ is continuous. I conduct inference at the three locations $v^*\in\{-0.25,0,0.25\}$, corresponding to three causal parameters: $\theta_0=\textsc{cate}(-0.25)=-0.10$, $\theta_0=\textsc{cate}(-0.25)=-0.10$, and $\theta_0=\textsc{cate}(0.25)=-0.32$. Figure~\ref{fig:design} visualizes $\textsc{cate}(v)$ and the three values of $\theta_0$. See Appendix~\ref{sec:details} for details on the data generating process.

\begin{wrapfigure}{R}{0.33\textwidth}
\vspace{-10pt}
	\centering
    \includegraphics[width=.3\textwidth]{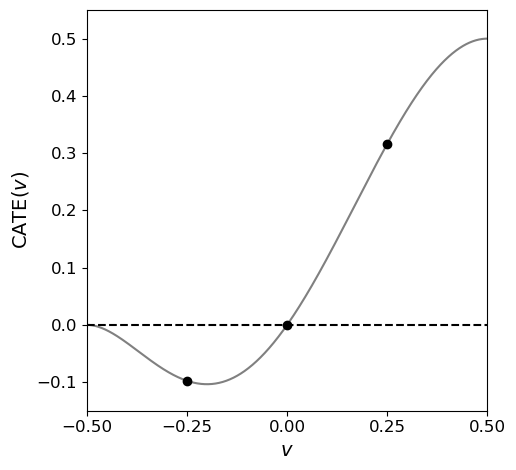}
    \vspace{-0pt}
    \caption{Simulation design. \textsc{cate}(v) is the curve and values of $\theta_0$ are the three points. 
    }
    \vspace{-10pt}
\label{fig:design}
\end{wrapfigure}

I evaluate confidence sets constructed from the KRRR estimator $\tilde{\alpha}$ (Definition~\ref{def:loss}) in the context of debiased machine learning $(\hat{\theta},\hat{\sigma}^2)$ (Definition~\ref{def:target}). I consider three variations of $(\hat{\theta},\hat{\sigma}^2)$, corresponding to different choices of the nonparametric regression estimator $\hat{\gamma}$: lasso, random forest, and neural network. Previous works on classical kernel ridge balancing weights typically disallow $\gamma_0\not\in H$, and appear not to justify inference on heterogeneous treatment effects. This use of kernel ridge balancing weights is justified by the main result of this paper. 

Following \cite[Tables S1 to S6]{chernozhukov2021simple}, I consider low dimensional and high dimensional specifications. In the low dimensional specification, the estimator $\hat{\gamma}$ uses $(D,X)$ as well as their interactions. In the high dimensional specification, $\hat{\gamma}$ uses fourth order polynomials. I focus here on the latter. See Appendix~\ref{sec:details} for the former.

Throughout, KRRR uses the Gaussian kernel for $X$, which corresponds to using all Hermite polynomials, appropriately weighted. 
See Appendix~\ref{sec:details} for implementation details.
 
An important choice is how to tune the bandwidth $h$ in $\textsc{local}\{(v-v^*)/h\}$. I follow the heuristic $h=c_h\hat{\sigma}_v n^{-1/5}$ of previous work, 
where $\hat{\sigma}^2_v$ is the sample variance of $V$. This heuristic satisfies the theoretical requirements $h=o(1)$ and $n^{-1/2}h^{-3/2}=o(1)$, and it implies a nonparametric rate of convergence of $(nh)^{-1/2}=n^{-2/5}$. The hyperparameter $c_h$ is chosen by the analyst. To evaluate robustness to tuning, I consider $c_h\in\{0.25, 0.50, 1.00\}$.

\begin{table*}[h]
\begin{centering}
\caption{High dimensional coverage simulations}
\label{tab:high}
\setlength{\tabcolsep}{7pt}
\renewcommand{\arraystretch}{0.7}
\resizebox{\textwidth}{!}{%
\begin{tabular}{cccccccccccc}
    \hline
    \multicolumn{3}{c}{} & \multicolumn{3}{c}{Lasso} & \multicolumn{3}{c}{Random Forest} & \multicolumn{3}{c}{Neural Network} \\
    \cmidrule(lr){4-6}\cmidrule(lr){7-9}\cmidrule(lr){10-12}
        & & & Ave. & Ave. & 95\% & Ave. & Ave. & 95\% & Ave. & Ave. & 95\% \\
    $v^*$ & \textsc{cate}($v^*$) & Tuning & Est. & S.E. & Cov. & Est. & S.E. & Cov. & Est. & S.E. & Cov. \\
    \hline
    -0.25 & -0.10 & 0.25 & -0.09 & 0.07 & 95\% & -0.10 & 0.05 & 93\% & -0.10 & 0.05 & 93\% \\
    -0.25 & -0.10 & 0.50 & -0.09 & 0.05 & 97\% & -0.09 & 0.04 & 95\% & -0.09 & 0.04 & 95\% \\
    -0.25 & -0.10 & 1.00 & -0.07 & 0.04 & 94\% & -0.08 & 0.02 & 82\% & -0.08 & 0.03 & 87\% \\
    0.00 & 0.00 & 0.25 & 0.02 & 0.06 & 96\% & 0.00 & 0.03 & 96\% & 0.00 & 0.03 & 97\% \\
    0.00 & 0.00 & 0.50 & 0.03 & 0.05 & 98\% & 0.01 & 0.02 & 94\% & 0.01 & 0.03 & 96\% \\
    0.00 & 0.00 & 1.00 & 0.05 & 0.04 & 87\% & 0.03 & 0.02 & 78\% & 0.03 & 0.02 & 80\% \\
    0.25 & 0.32 & 0.25 & 0.38 & 0.17 & 97\% & 0.36 & 0.16 & 95\% & 0.34 & 0.14 & 96\% \\
    0.25 & 0.32 & 0.50 & 0.37 & 0.12 & 99\% & 0.35 & 0.11 & 95\% & 0.34 & 0.10 & 97\% \\
    0.25 & 0.32 & 1.00 & 0.35 & 0.09 & 99\% & 0.34 & 0.09 & 96\% & 0.32 & 0.08 & 97\% \\
    \hline
\end{tabular}
}
\end{centering}
\end{table*}

Tables~\ref{tab:high} and~\ref{tab:low} present results across 500 simulations. The initial columns denote the choice of $v^*$ and true value $\theta_0=\textsc{cate}(v^*)$. The next column is the hyperparameter value $c_h$. I report the average point estimate, average standard error, and 95\% coverage across 500 simulations for a given choice of $(v^*,\theta_0,c_h)$. 

Coverage is close to nominal and quite stable when $c_h\in\{0.25, 0.5\}$, across estimators and across locations $v^*$. When $c_h$ is too large, e.g. $c_h=1.00$, there is some under coverage for the random forest and neural network for the locations $v^*\in\{-0.25,0.00\}$. 

Compared to results for lasso Riesz representers (LRR) \cite{chernozhukov2018learning}, KRRR considerably reduces the absolute bias in this smooth design. See \cite[Tables S4-S6]{chernozhukov2021simple} for values analogous to those in Table~\ref{tab:high}. 
In particular at the location $v^*=0.25$, for lasso regression, KRRR gives absolute bias $(0.06,0.05,0.03)$ while LRR gives absolute bias $(0.09,0.13,0.11)$;
for random forest regression, KRRR gives $(0.04,0.04,0.02)$ while LRR gives $(0.05,0.08,0.07)$; 
and for neural network regression, KRRR gives $(0.02,0.02,0.00)$ while LRR gives $(0.07,0.07,0.06)$.

\subsection{Confidence sets for heterogeneous effects of 401(k)}

Having justified the pointwise coverage of KRRR for heterogeneous treatment effects, I now use the procedure to quantify uncertainty for the heterogeneous effects of 401(k) eligibility on assets by age. The empirical analysis yields meaningful economic insights.

I follow the identification strategy of \cite{poterba1995}. The authors assume that when
401(k) was introduced, workers ignored whether jobs offered 401(k) plans and made employment decisions based on income and other observable job characteristics. They assume 401(k) eligibility was as good as randomly assigned, conditional on covariates.

I use data from the 1991 US Survey of Income and Program Participation, following the sample selection and variable construction of \cite{chernozhukov2004effects}. The outcome $Y$ is net financial
assets defined as the sum of IRA balances, 401(k) balances, checking accounts, US saving bonds,
other interest-earning accounts, stocks, mutual funds, and other interest-earning assets minus nonmortgage debt. The treatment $D$ is eligibility
to enroll in a 401(k) plan. The covariates $X$ are age, income, years of education, family size, marital
status, two-earner status, benefit pension status, IRA participation, and home-ownership. The data include $n = 9915$ observations. 

As in the simulations, I consider a high dimensional specification and various choices of the nonparametric regression estimator $\hat{\gamma}$: lasso, random forest, and neural network. I use KRRR $\tilde{\alpha}$ with the Gaussian kernel. See Appendix~\ref{sec:details} for the low dimensional specification.

\begin{figure}
	\centering
    \subfloat[Lasso]{%
    \includegraphics[height=0.21\textheight]{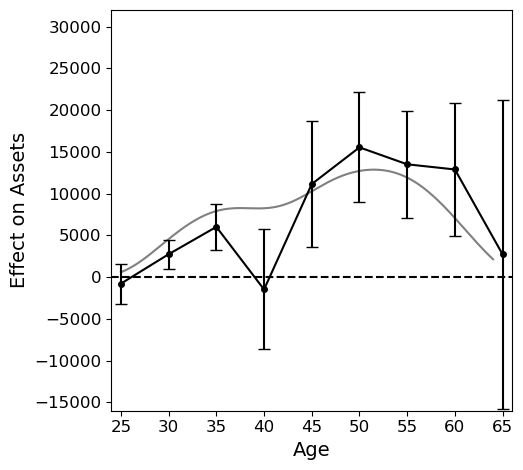}}
    \subfloat[Random forest]{%
    \includegraphics[height=0.21\textheight]{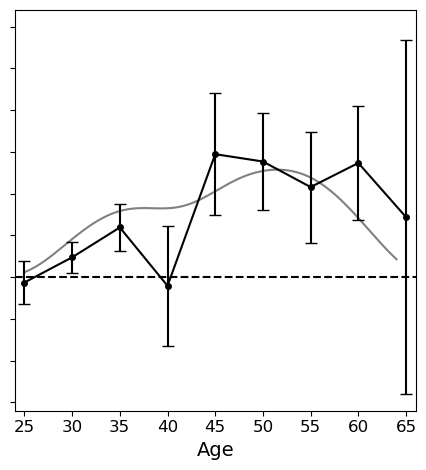}}
    \subfloat[Neural network]{%
    \includegraphics[height=0.21\textheight]{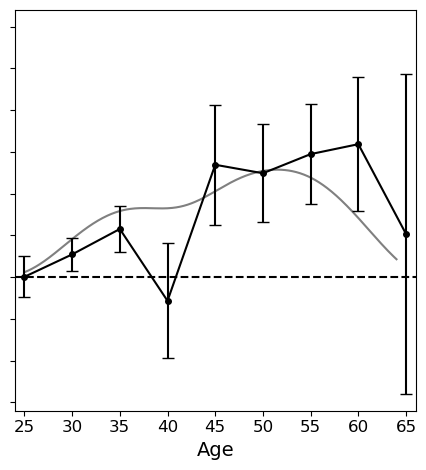}}
    \vspace{-10pt}
    \caption{Heterogeneous treatment effects by age. The point estimates and confidence sets use KRRR.
    The smooth curve is a comparison estimator \cite{singh2020kernel}. 
    }
\label{fig:401_high}
\end{figure}

Figures~\ref{fig:401_high} and~\ref{fig:401_low} visualize point estimates and pointwise 95\% confidence sets based on KRRR, in black. I find that 401(k) eligibility has positive and statistically significant effects that vary by age. The effects appear to be small and positive for 30 year olds (about \$2,500), and generally increasing, before plateauing for 45 to 60 year olds (about \$12,500). The effect is not statistically significant for 65 year olds. 

In summary, under the stated assumptions, 401(k) eligibility seems to cause middle aged individuals to save about five times more than 30 years olds. Individuals in the former subpopulation tend to be at the peak of their earning potentials, while those in the latter subpopulation are closer to the beginning of their careers. The results are robust, across variations of the regression estimator and the specification.

For comparison, I also visualize the smooth nonparametric estimator of \cite{singh2020kernel}, in grey. The confidence sets of this work corroborate their estimate, and quantify uncertainty. In particular, this paper's confidence sets suggest that the apparent dip in effects for 65 year olds is not statistically significant.

\section{Discussion}\label{sec:conclusion}

Causal inference introduces a counterfactual effective dimension. Under standard regularity conditions on causal parameters from semiparametric theory, e.g. a propensity score bounded away from zero and one, I show that the counterfactual effective dimension is not much greater than the actual effective dimension. Using this technique, I prove population $L_2$ rates of generalization error for kernel ridge balancing weights, similar to kernel ridge regression. These rates connect kernel ridge balancing weights with general semiparametric results which tolerate $\gamma_0\not \in H$ and justify confidence sets for causal functions. 

Future work may prove a comprehensive lower bound on $L_2$ rates for Riesz representers in the RKHS framework, and evaluate whether KRRR rates are optimal in general. Future work may also study benign overfitting for KRRR, similar to kernel ridge regression, using the counterfactual effective dimension technique developed in this paper.

\spacingset{1}

\bibliographystyle{apalike}
\DeclareRobustCommand{\VAN}[2]{#2}  
\DeclareRobustCommand{\VAN}[2]{#1}  

\newpage 

\spacingset{1.65}


\appendix
\section{Formal inference on causal functions}\label{sec:nonparametric}

\subsection{Framework}

By formal inference, I mean justification for Gaussian approximation and nominal coverage guarantees.
In this section, I provide further details for the pointwise approximation of nonparametric causal functions. I discuss parameters of the form
$$
\theta_0^{\lim}=\lim_{h\rightarrow 0} \theta_0^h, \quad \theta_0^h=\mathbb{E}\{m_h(W,\gamma_0)\}=\mathbb{E}\{\ell_h(W)\check{m}(W,\gamma_0)\}
$$
where $\ell_h(w)$ is a Nadaraya Watson weighting with bandwidth $h$ for a scalar component of $W$. While $\theta_0^{\lim}$ is a nonparametric object, it may be approximated by the sequence $(\theta_0^h)$ of semiparametric objects as long as we keep track of how certain quantities depend on $h$. The approximation error is $\Delta_h=n^{1/2}\sigma_h^{-1}|\theta_0^h-\theta_0^{\lim}|$.

Recall Example~\ref{ex:het}. Here, $W=(D,X)$ concatenates the treatment $D\in\{0,1\}$ and covariates $X$, and we wish to study heterogeneity with respect to the first covariate $X_1\in\mathbb{R}$. Let $x_1^*$ be a particular value of $X_1$. Then the heterogeneous treatment effect for the subpopulation with $X_1=x_1^*$ is $\theta_0^{\lim}=\lim_{h\rightarrow 0} \theta_0^h$ where
$
\theta_0^h=\mathbb{E}\{\ell_h(W)\{\gamma_0(1,X)-\gamma_0(0,X)\}$, 
$$
\ell_h(W)=\textsc{local}\{(X_1-x_1^*)/h\}=\frac{\mathcal{K}\{(X_1-x_1^*)/h\}}{\mathbb{E}[\mathcal{K}\{(X_1-x_1^*)/h\}]},
$$
and $\mathcal{K}$ is a bounded and symmetric kernel that integrates to one.

\subsection{Technical details}

It is convenient to refer to the Riesz representer of $f\mapsto \mathbb{E}\{m_h(W,f)\}$ by $\alpha_h$ and the Riesz representer of $f\mapsto \mathbb{E}\{\check{m}(W,f)\}$ by $\check{\alpha}_0$. In Example~\ref{ex:het}, 
$$
\alpha_h(W)=\ell_h(W) \check{\alpha}_0(W),\quad \check{\alpha}_0(W)=\frac{D}{\pi_0(D)}-\frac{1-D}{1-\pi_0(D)},\quad \pi_0(X)=\mathbb{E}(D|X)
$$
and $\ell_h(W)$ is given above.
For simplicity, in this discussion, I set aside mis-specification; $\gamma_{\mathcal{G}}=\gamma_0$ and $\check{\alpha}_{\mathcal{A}}=\check{\alpha}_0$. It can be re-introduced in a straightforward manner.

Lemma~\ref{lemma:dml} in the main text holds for $m_h$ and $\alpha_h$. In particular the rate conditions are for $\mathcal{R}(\hat{\alpha}_h)$, i.e. the rate at which $\hat{\alpha}_h$ converges to $\alpha_h$.

I now clarify how various quantities in Lemma~\ref{lemma:dml} depend on $h$. Consider the estimator $\hat{\alpha}_h(W)=\ell_h(W)\tilde{\alpha}(W)$, where $\tilde{\alpha}$ is the KRRR estimator for $\check{\alpha}_0$. Section~\ref{sec:experiments} confirms that this choice performs well in simulations.  The results below are in terms of $h$ and $\mathcal{R}(\tilde{\alpha})=\mathbb{E}[\{\tilde{\alpha}(W)-\check{\alpha}_0(W)\}^2|(W_i)_{i\in[n]}]$, i.e. the rate at which $\tilde{\alpha}$ converges to $\check{\alpha}_0$.

\begin{lemma}[Characterization of key quantities; Theorem 2 of \cite{chernozhukov2021simple}]\label{lemma:local}
If noise has finite variance then $\bar{\sigma}^2<\infty$. 
Suppose bounded moment, heteroscedasticity, density, and derivative conditions of \cite[Supplement 2.4]{chernozhukov2021simple} hold. Then the moments satisfy
$$
\zeta_h/\sigma_h \lesssim h^{-1/6},\quad \sigma_h \asymp \bar{M}_h \asymp h^{-1/2},\quad \zeta_h\lesssim h^{-2/3},\quad \chi_h\lesssim h^{-3/4}.$$
The other key quantities in Lemma~\ref{lemma:dml} satisfy
$$
\bar{M}_h\lesssim h^{-2},\quad \bar{\alpha}_h \lesssim h^{-1},\quad \mathcal{R}(\hat{\alpha}_h)\lesssim h^{-2}\mathcal{R}(\tilde{\alpha}),\quad \Delta_h \lesssim n^{1/2} h^{\mathsf{v}+1/2}
$$
where $\mathsf{v}$ is the order of differentiability of $\mathcal{K}$.
\end{lemma}

\begin{corollary}[Inference for causal functions; Corollary S1 of \cite{chernozhukov2021simple}]
    Suppose the conditions of Lemmas~\ref{lemma:dml} and~\ref{lemma:local} hold as well as correct specification. As $n\rightarrow \infty$, suppose the bandwidth satisfies $h=o(1)$ and $n^{-1/2}h^{-3/2}=o(1)$. Suppose the following learning rate conditions hold:
\begin{enumerate}
    \item $\left(h^{-1}+\bar{\alpha}'\right)\mathcal{R}(\hat{\gamma})^{1/2}=o_p(1)$;
    \item $\bar{\sigma}h^{-1}\mathcal{R}(\tilde{\alpha})^{1/2}=o_p(1)$;
    \item $h^{-1/2}\{n \mathcal{R}(\hat{\gamma}) \mathcal{R}(\tilde{\alpha})\}^{1/2} =o_p(1)$.
\end{enumerate}
Finally suppose the approximation error condition $\Delta_h \rightarrow 0$. Then $\sigma_h^{-1}n^{1/2}(\hat{\theta}^h-\theta^{\lim}_0)\leadsto \mathcal{N}(0,1)$ and $\mathbb{P} \left\{\theta^{\lim}_0 \in \left(\hat{\theta}^h\pm c_a\hat{\sigma} n^{-1/2} \right)\right\}\rightarrow 1-a$.
\end{corollary}
\section{Deferred proofs from Section~\ref{sec:algorithm}}\label{sec:algorithm_supp}

\subsection{General loss function}

\begin{proof}[Proof of Lemma~\ref{lemma:loss}]
To begin, write 
$$\|\alpha-\alpha_0\|^2_2=\|\alpha_0\|_2^2-2\mathbb{E}\{\alpha_0(W)\alpha(W)\}+\|\alpha\|_2^2.$$ 
The first term does not depend on $\alpha$. Kernel boundedness in Assumption~\ref{assumption:RKHS} implies Bochner integrability \cite[Definition A.5.20]{steinwart2008support}, which allows us to exchange expectation and RKHS inner product in the second and third terms. By Lemma~\ref{lemma:riesz}, the definition of counterfactual features, and boundedness of the counterfactual kernel, the second term has
$$
    \mathbb{E}\{\alpha_0(W)\alpha(W)\}=\mathbb{E}\{m(W,\alpha)\}
    =\mathbb{E}\{\langle \alpha,\phi^{(m)}(w) \rangle_H\}
    =\langle \alpha, \mu^{(m)} \rangle_H.
$$
By the reproducing property and boundedness of the actual kernel, the third term is
$$
\|\alpha\|_2^2=\mathbb{E}\{\alpha(W)^2\}]=\mathbb{E}\{\langle \alpha,\phi(W)\rangle^2_H\}
=\mathbb{E}[\langle \alpha,\{\phi(W) \otimes \phi(W)^*\} \alpha \rangle_H]
=\langle \alpha, T\alpha\rangle_H.
$$
\end{proof}

\begin{proof}[Proof of Lemma~\ref{lemma:foc}]
    Take derivatives of the various losses in Definitions~\ref{def:approx} and~\ref{def:loss}. See \cite[Proof of Proposition 2]{de2005risk} for technical details of differentiation in the RKHS. 
\end{proof}

\begin{proof}[Proof of Corollary~\ref{cor:trade}]
    Rearranging Lemma~\ref{lemma:foc} gives $\hat{T}\tilde{\alpha}-\hat{\mu}^{(m)}=-\lambda \tilde{\alpha}$. Taking the RKHS norm of both sides, $\|\hat{T}\tilde{\alpha}-\hat{\mu}^{(m)}\|_H=\lambda \|\tilde{\alpha}\|_H$. Finally, write 
    $$\hat{T}\tilde{\alpha}
    =\mathbb{E}_n[\{\phi(W)\otimes \phi(W)^*\}\tilde{\alpha}]
    =\mathbb{E}_n\{\phi(W)\langle \phi(W),\tilde{\alpha}\rangle_{H}\}
    =\mathbb{E}_n\{\phi(W)\tilde{\alpha}(W)\}$$ and $\hat{\mu}^{(m)}=\mathbb{E}_n\{\phi^{(m)}(W)\}$.
\end{proof}

\subsection{Detailed comparisons to kernel ridge regression and kernel ridge balancing weights}

\begin{proof}[Proof of Corollary~\ref{cor:krr}]
   The standard formula for ridge regression is $\tilde{\gamma}=(\hat{T}+\lambda)^{-1}\mathbb{E}_n\{Y\phi(W)\}$. Comparing to the formula $\tilde{\alpha}=(\hat{T}+\lambda)^{-1}\hat{\mu}^{(m)}$ in Lemma~\ref{lemma:foc}, $\mathbb{E}_n\{Y\phi(W)\}$ equals $\hat{\mu}^{(m)}=\mathbb{E}\{\phi^{(m)}(W)\}$ when $m(w,f)=yf(w)$.
\end{proof}

\begin{proof}[Proof of Corollary~\ref{cor:hirshberg}]
I restate the proof of \cite[p. 5]{bruns2023augmented} for completeness. Define
$$\textsc{imbalance}(\beta)=\sup_{f\in H:\|f\|_H\leq 1}\frac{1}{n}\sum_{i=1}^n \beta_i f(W_i)-m(W_i,f)$$ so that $\tilde{\mathcal{L}}_{\lambda}^n(\beta)=\{\textsc{imbalance}(\beta)\}^2+\lambda n^{-1}\beta^{\top}\beta$. 

To begin, use the reproducing property to write
$$
\beta_i f(W_i)-m(W_i,f)
=\beta_i \langle f,\phi(W_i)\rangle_H-\langle f,\phi^{(m)}(W_i)\rangle_H
=\langle f,\beta_i\phi(W_i)-\phi^{(m)}(W_i) \rangle_H.
$$
Since $H$ is self-dual,
\begin{align*} 
\textsc{imbalance}(\beta)
&=\sup_{f\in H:\|f\|_H\leq 1}\frac{1}{n}\sum_{i=1}^n \langle f,\beta_i\phi(W_i)-\phi^{(m)}(W_i) \rangle_H \\
&=
\sup_{f\in H:\|f\|_H\leq 1}\left\langle f,\frac{1}{n}\sum_{i=1}^n \{\beta_i\phi(W_i)-\phi^{(m)}(W_i)\} \right\rangle_H \\
&=\left\|\frac{1}{n}\sum_{i=1}^n \{\beta_i\phi(W_i)-\phi^{(m)}(W_i)\}\right\|_H
=\|\Phi^*\beta/n-\hat{\mu}^{(m)}\|_H
\end{align*}
where $\Phi^*$ is the adjoint of the operator $\Phi:H\rightarrow  \mathbb{R}^n$ whose $i$th component is $\langle \phi(W_i),\cdot\rangle_H$. 

Therefore the loss $ \tilde{\mathcal{L}}_{\lambda}^n(\beta)$ becomes 
\begin{align*}
   \|\Phi^*\beta/n-\hat{\mu}^{(m)}\|^2_H+\lambda n^{-1}\beta^{\top}\beta
    =\frac{1}{n^2}\beta^{\top} \Phi \Phi^*\beta -\frac{2}{n}\beta^{\top}\Phi \hat{\mu}^{(m)} +\|\hat{\mu}^{(m)}\|_H^2+\frac{\lambda}{n}\beta^{\top}\beta.
\end{align*}
Its first order condition yields
$\tilde{\beta}=(\Phi\Phi^*/n+\lambda)^{-1}\Phi\hat{\mu}^{(m)}$ whose $i$th component is $\beta_i$. 

By Lemma~\ref{lemma:foc}, $\Phi\tilde{\alpha}=\Phi(\Phi^*\Phi/n+\lambda)^{-1}\hat{\mu}^{(m)}$ whose $i$th component is $\langle \phi(W_i),\tilde{\alpha}\rangle_H=\tilde{\alpha}(W_i)$. Thus what remains is to show $(\Phi\Phi^*/n+\lambda)^{-1}\Phi=\Phi(\Phi^*\Phi/n+\lambda)^{-1}$, which follows by expressing $\Phi$ via its singular value decomposition $U\Sigma V^*$. In particular, 
\begin{align*}
    (\Phi\Phi^*/n+\lambda)^{-1}\Phi
&
=U(\Sigma^2n^{-1}+\lambda)^{-1} U^*U\Sigma V^*
=U(\Sigma^2n^{-1}+\lambda)^{-1}\Sigma V^* \\
\Phi(\Phi^*\Phi/n+\lambda)^{-1}
&
=U\Sigma V^*V(\Sigma^2 n^{-1}+\lambda)^{-1} V^*
=U\Sigma (\Sigma^2 n^{-1}+\lambda)^{-1} V^*
\end{align*}
and $(\Sigma^2n^{-1}+\lambda)^{-1}\Sigma=\Sigma (\Sigma^2 n^{-1}+\lambda)^{-1}$ by diagonality of each factor.
\end{proof}

\begin{proof}[Proof of Corollary~\ref{cor:equal}]
   By the reproducing property, boundedness of the counterfactual kernel, and the well known expression for kernel ridge regression, write 
   $$
   \mathbb{E}_n\{m(W,\tilde{\gamma})\}
   =\mathbb{E}_n\{\langle \tilde{\gamma},\phi^{(m)}(W)\rangle_H\}
   =\langle \tilde{\gamma},\hat{\mu}^{(m)}\rangle_H
   =\langle (\hat{T}+\lambda)^{-1}\mathbb{E}_n\{Y\phi(W)\},\hat{\mu}^{(m)}\rangle_H.
   $$   
   By the reproducing property, boundedness of the actual kernel, and Lemma~\ref{lemma:foc}, write 
   \begin{align*}
        \mathbb{E}_n\{Y\tilde{\alpha}(W)\}
   &=\mathbb{E}_n\{Y\langle \tilde{\alpha},\phi(W)\rangle_H\}
   =\langle \tilde{\alpha},\mathbb{E}_n\{Y\phi(W)\}\rangle_H\\
   &=\langle (\hat{T}+\lambda)^{-1}\hat{\mu}^{(m)},\mathbb{E}_n\{Y\phi(W)\}\rangle_H.
   \end{align*}
   Finally observe that $(\hat{T}+\lambda)^{-1}$ is self adjoint.
\end{proof}

\begin{proof}[Proof of Corollary~\ref{cor:non-equal}]
    See the proof of Corollary~\ref{cor:equal}.
\end{proof}

\subsection{Standalone closed form solution}

\begin{proof}[Proof of Lemma~\ref{lemma:represent}]
The argument is based on \cite[Lemma 2]{chernozhukov2020adversarial}, which instead finds a solution in $\mathbb{R}^{2n}$ for an adversary that tries to violate Riesz representation.

By Definition~\ref{def:loss}, the definitions of $\hat{\mu}^{(m)}$ and $\hat{T}$, and boundedness of the kernels,
\begin{align*}
\mathcal{L}^n_{\lambda}(\alpha)
&=-2\langle \alpha, \hat{\mu}^{(m)} \rangle_H+\langle \alpha, \hat{T}\alpha\rangle_H+\lambda \|\alpha\|^2_H\\
&=-2\langle \alpha, \mathbb{E}_n\{\phi^{(m)}(W)\} \rangle_H+\langle \alpha, \mathbb{E}_n\{\phi(W)\otimes \phi(W)^*\}\alpha\rangle_H+\lambda \|\alpha\|^2_H \\
&=\mathbb{E}_n[-2\langle \alpha, \phi^{(m)}(W) \rangle_H+\langle \alpha, \{\phi(W)\otimes \phi(W)^*\}\alpha\rangle_H]+\lambda \|\alpha\|^2_H \\
&=\mathbb{E}_n\{-2\langle \alpha, \phi^{(m)}(W) \rangle_H+\langle \alpha, \phi(W)\rangle_H \langle \alpha, \phi(W)\rangle_H \}+\lambda \|\alpha\|^2_H.
\end{align*}
Due to the ridge penalty, $\mathcal{L}^n_{\lambda}$ is coercive and strongly convex. Hence it has a unique minimizer $\tilde{\alpha}$.
Write $\tilde{\alpha}=\tilde{\alpha}_n+\tilde{\alpha}^{\perp}_n$ where $\tilde{\alpha}_n\in row(\Psi)$ and $\tilde{\alpha}_n^{\perp}\in null(\Psi)$. In words, $\tilde{\alpha}^{\perp}_n$ is orthogonal to the features and counterfactual features evaluated at data. I use this orthogonality and the final expression in the display to conclude that
$
\mathcal{L}^n_{\lambda}(\tilde{\alpha})=\mathcal{L}^n_{\lambda}(\tilde{\alpha}_n)+\lambda \|\tilde{\alpha}_n^{\perp}\|^2_H.
$
Therefore
$
\mathcal{L}^n_{\lambda}(\tilde{\alpha})\geq \mathcal{L}^n_{\lambda}(\tilde{\alpha}_n)
$.
Since $\tilde{\alpha}$ is the unique minimizer, $\tilde{\alpha}=\tilde{\alpha}_n$.
\end{proof}

\begin{proof}[Proof of Propositon~\ref{prop:closed}]
    Using Definition~\ref{def:loss} and Lemma~\ref{lemma:represent}, we have
$$
\mathcal{L}^n_{\lambda}(\tilde{\alpha})=-2\langle \tilde{\alpha}, \hat{\mu}^{(m)} \rangle_H+\langle \tilde{\alpha}, \hat{T}\tilde{\alpha}\rangle_H+\lambda \|\alpha\|^2_H=-2\rho^\top \Psi \hat{\mu}^{(m)}+\rho^{\top}\Psi\hat{T}\Psi^*\rho+\lambda \rho^\top \Psi \Psi^* \rho.$$ 
The first order condition yields 
$$
\hat{\rho}=\{\Psi\hat{T}\Psi^*+\lambda\Psi\Psi^*\}^{-1}\Psi\hat{\mu}^{(m)}=\{\Psi\Phi^*\Phi \Psi^*+n\lambda\Psi\Psi^*\}^{-1}\{n \Psi \hat{\mu}^{(m)}\}.$$ 
To evaluate the estimator at a test location $w$, compute
$
\tilde{\alpha}(w)=\langle \tilde{\alpha}, \phi(w)\rangle_H =\rho^{\top}\Psi\phi(w).
$
Finally, note that $\Omega=\Psi \Phi^*\Phi \Psi^*$, $K=\Psi \Psi^*$, $v=n\Psi\hat{\mu}^{(m)}$, and $u(w)=\Psi\phi(w)$. 
\end{proof}
\section{Deferred proofs from Section~\ref{sec:mse}}\label{sec:mse_details}

\subsection{Main result}

\begin{lemma}[Standard learning theory bounds; Proposition 3 of \cite{caponnetto2007optimal}]\label{lemma:prior}
Under Assumption~\ref{assumption:c},
$
    \mathcal{A}(\lambda)\leq \lambda^cR $ and $
    \mathcal{B}(\lambda)\leq \lambda^{c-1}R$, where $R=\|T^{\frac{1-c}{2}}\alpha_H\|^2_H$.
Under Assumption~\ref{assumption:b}(i),
$
\mathcal{N}(\lambda)\leq \bar{B}^{\frac{1}{b}}\frac{\pi/b}{\sin(\pi/b)}\lambda^{-\frac{1}{b}}$ if $b<\infty$, and  $
\mathcal{N}(\lambda)\leq J$ if $b=\infty$ .
\end{lemma}

\begin{proof}
    Both $ \mathcal{A}(\lambda)$ and $ \mathcal{B}(\lambda)$ are standard. See \cite[Section 1]{sutherland2017fixing} for $\mathcal{N}(\lambda)$.
\end{proof}

\begin{proof}[Proof of Theorem~\ref{theorem:riesz}]
I combine the nonasymptotic bound in Proposition~\ref{prop:abstract} with the learning theory bounds of Lemma~\ref{lemma:prior}, following \cite[Section 2]{sutherland2017fixing}. The absolute constant depends only on the quantities $(R,\kappa,\kappa^{(m)},\Upsilon,\Sigma,\underline{B}, \bar{B},b,c)$. Note that $(\Upsilon,\Sigma)$ introduce dependence on $(\|\alpha_H\|_H,\bar{M})$. Finally appeal to Lemma~\ref{lemma:norms}.
\end{proof}

\begin{proof}[Proof of Corollary~\ref{cor:cap}]
    By Corollary~\ref{cor:krr}, KRRR nests kernel ridge regression. If $|Y|\leq \bar{Y}$, then Assumption~\ref{assumption:cont} holds with $\bar{M}=\bar{Y}^2$. Finally appeal to Theorem~\ref{theorem:riesz}.
\end{proof}

\begin{proof}[Proof of Corollary~\ref{cor:optimal}]
    The conclusion is immediate.
\end{proof}

\subsection{Semiparametric inference beyond the RKHS}

Train $(\hat{\gamma}_{\ell},\hat{\alpha}_{\ell})$ on observations in $I_{\ell}^c$. 
Let $n_{\ell}=|I_{\ell}|=n/L$ be the number of observations in $I_{\ell}$. 
Denote by $\mathbb{E}_{\ell}(\cdot)=n_{\ell}^{-1}\sum_{i\in I_{\ell}}(\cdot)$ the average over observations in $I_{\ell}$. Denote by $\mathbb{E}_n(\cdot)=n^{-1}\sum_{i=1}^n(\cdot)$ the average over all observations in the sample.

To study mis-specification, let $\gamma_{\mathcal{G}}$ and $\alpha_{\mathcal{A}}$ be the best in class approximations to $\gamma_0$ and $\alpha_0$, respectively. I reduce the error bound to rates at which $\hat{\gamma}\in\mathcal{G}$ converges to $\gamma_{\mathcal{G}}$, and $\hat{\alpha}\in\mathcal{A}$ converges to $\alpha_{\mathcal{A}}$, respectively.

\begin{definition}[Foldwise target and oracle]
\begin{align*}
 \hat{\theta}_{\ell}&=\mathbb{E}_{\ell} [m(W,\hat{\gamma}_{\ell})+\hat{\alpha}_{\ell}(W)\{Y-\hat{\gamma}_{\ell}(W)\}],\;
    \bar{\theta}_{\ell}=\mathbb{E}_{\ell} [m(W,\gamma_{\mathcal{G}})+\alpha_{\mathcal{A}}(W)\{Y-\gamma_{\mathcal{G}}(W)\}].
\end{align*}
\end{definition}

\begin{definition}[Overall target and oracle]
$$
    \hat{\theta}=\frac{1}{L}\sum_{\ell=1}^L \hat{\theta}_{\ell},\quad 
    \bar{\theta}=\frac{1}{L}\sum_{\ell=1}^L \bar{\theta}_{\ell},\quad \theta_{\textsc{mis}}=\mathbb{E} [m(W,\gamma_{\mathcal{G}})+\alpha_{\mathcal{A}}(W)\{Y-\gamma_{\mathcal{G}}(W)\}].
$$
\end{definition}

\begin{lemma}[Residual expansion]\label{lemma:Taylor}
Let $u=\hat{\gamma}_{\ell}-\gamma_{\mathcal{G}}$ and $v=\hat{\alpha}_{\ell}-\alpha_{\mathcal{A}}$. Then $|\hat{\theta}_{\ell}-\bar{\theta}_{\ell}|\leq \sum_{j=1}^3 \Delta_{j{\ell}}^{1/2}$ where
\begin{align*}
    \Delta_{1{\ell}}&=\mathbb{E}_{\ell}\{m(W,u)^2\},\quad 
    \Delta_{2{\ell}}=\mathbb{E}_{\ell}\{u(W)^2\hat{\alpha}_{\ell}(W)^2\},\quad
    \Delta_{3{\ell}}=\mathbb{E}_{\ell}[v(W)^2\{Y-\gamma_{\mathcal{G}}(W)\}^2].
\end{align*}
\end{lemma}

\begin{proof}
Write the difference within the empirical mean as
\begin{align*}
    &m(W,\hat{\gamma}_{\ell})+\hat{\alpha}_{\ell}(W)\{Y-\hat{\gamma}_{\ell}(W)\}-[m(W,\gamma_{\mathcal{G}})+\alpha_{\mathcal{A}}(W)\{Y-\gamma_{\mathcal{G}}(W)\}] \\
    &=m(W,\hat{\gamma}_{\ell}-\gamma_{\mathcal{G}}) +\hat{\alpha}_{\ell}(W)\{Y-\hat{\gamma}_{\ell}(W)\} - \hat{\alpha}_{\ell}(W)\{Y-\gamma_{\mathcal{G}}(W)\} \\
    &\quad + \hat{\alpha}_{\ell}(W)\{Y-\gamma_{\mathcal{G}}(W)\} -\alpha_{\mathcal{A}}(W)\{Y-\gamma_{\mathcal{G}}(W)\} \\
    &=m(W,\hat{\gamma}_{\ell}-\gamma_{\mathcal{G}}) 
    +\hat{\alpha}_{\ell}(W)\{\gamma_{\mathcal{G}}-\hat{\gamma}_{\ell}(W)\}
    +\{\hat{\alpha}_{\ell}(W)-\alpha_{\mathcal{A}}(W)\}\{Y-\gamma_{\mathcal{G}}(W)\}.
\end{align*}
Averaging over observations in $I_{\ell}$,
$$
\hat{\theta}_{\ell}-\bar{\theta}_{\ell}=\mathbb{E}_{\ell}\{m(W,u)\}+\mathbb{E}_{\ell}\{-u(W)\hat{\alpha}_{\ell}(W)\}+\mathbb{E}_{\ell}[v(W)\{Y-\gamma_{\mathcal{G}}(W)\}].
$$
Finally, apply Jensen's inequality to each term.
\end{proof}

\begin{lemma}[Residuals; Proposition S9 of \cite{chernozhukov2021simple}]\label{lemma:resid2}
Suppose Assumption~\ref{assumption:cont} holds and
$$
\mathbb{E}[\{Y-\gamma_{\mathcal{G}}(W)\} \mid W ]^2\leq \bar{\sigma}^2,\quad \|\hat{\alpha}_{\ell}\|_{\infty}\leq\bar{\alpha}'.
$$
Then with probability $1-\epsilon/(2L)$, for a fixed $\ell \in [L]$, jointly
\begin{align*}
    \Delta_{1\ell}&\leq t_1=\frac{6L}{\epsilon}\bar{M}\mathcal{R}(\hat{\gamma}_{\ell}),\quad
     \Delta_{2\ell}\leq t_2=\frac{6L}{\epsilon}(\bar{\alpha}')^2\mathcal{R}(\hat{\gamma}_{\ell}),\quad
     \Delta_{3\ell}\leq t_3=\frac{6L}{\epsilon}\bar{\sigma}^2\mathcal{R}(\hat{\alpha}_{\ell}). 
\end{align*}
\end{lemma}

\begin{definition}[Shorter notation]
Define the notation
$$
 \psi_{\textsc{mis}}(w)=\psi(w,\theta_0,\gamma_{\mathcal{G}},\alpha_{\mathcal{A}}), \quad \psi(w,\theta,\gamma,\alpha)=m(w,\gamma)+\alpha(w)\{y-\gamma(w)\}-\theta,
$$
where $\gamma\mapsto m(w,\gamma)$ is linear. Let $\sigma_{\textsc{mis}}^2=var\{\psi_{\textsc{mis}}(W)\}$.
\end{definition}

\begin{lemma}[Oracle concentration]\label{lemma:oracle}
    Suppose $\sigma_{\textsc{mis}}^2<\infty$. Then with probability $1-\epsilon/2$,
    $$
   |\bar{\theta}-\theta_{\textsc{mis}}|=\Delta \leq t=\left(\frac{2}{\epsilon}\right)^{1/2}\frac{\sigma_{\textsc{mis}}}{n^{1/2}}.
    $$
\end{lemma}

\begin{proof}
I proceed in steps.
    \begin{enumerate}
        \item Decomposition. Write
      \begin{align*}
        &\bar{\theta}-\theta_{\textsc{mis}} \\
        &=\mathbb{E}_n [m(W,\gamma_{\mathcal{G}})+\alpha_{\mathcal{A}}(W)\{Y-\gamma_{\mathcal{G}}(W)\}]-\mathbb{E}[m(W,\gamma_{\mathcal{G}})+\alpha_{\mathcal{A}}(W)\{Y-\gamma_{\mathcal{G}}(W)\}] \\
        &=\mathbb{E}_n \{\psi_{\textsc{mis}}(W)\}-\mathbb{E}\{\psi_{\textsc{mis}}(W)\}=\mathbb{E}_n\{\xi(W)\}
       \end{align*}
       where $\xi(W)=\psi_{\textsc{mis}}(W)-\mathbb{E}\{\psi_{\textsc{mis}}(W)\}$ is mean zero by construction.
        \item Bounding moments. Since observations are independent and identically distributed,
        \begin{align*}
            \mathbb{E}(\Delta^2)
            &=\mathbb{E} ([\mathbb{E}_n\{\xi(W)\}]^2)
            =\mathbb{E} \left\{\frac{1}{n^2} \sum_{i,j\in [n]} \xi(W_i)\xi(W_j)\right\} \\
            &=\frac{1}{n^2} \sum_{i,j\in [n]} \mathbb{E} \{\xi(W_i)\xi(W_j)\} 
            =\frac{1}{n^2} \sum_{i\in [n]} \mathbb{E} \{\xi(W_i)^2\} 
            =\frac{1}{n}\mathbb{E} \{\xi(W)^2\}=\frac{1}{n}\sigma_{\textsc{mis}}^2. 
        \end{align*}
       \item Markov inequality implies
        $$
        \mathbb{P}(\Delta>t)\leq \frac{\mathbb{E}(\Delta^2)}{t^2}= \frac{\sigma_{\textsc{mis}}^2}{nt^2} =\frac{\epsilon}{2}.
        $$ 
        Solving for $t$ gives the desired result.
    \end{enumerate}
\end{proof}

\begin{proposition}[Nonasymptotic bound under mis-specification]\label{prop:non}
    Suppose the conditions of Lemmas~\ref{lemma:resid2} and~\ref{lemma:oracle} hold. Then with probability $1-\epsilon$,
    $$
    |\hat{\theta}-\theta_{\textsc{mis}}|\leq \left(\frac{6L}{\epsilon}\right)^{1/2}\left\{
(\bar{M}^{1/2}+\bar{\alpha}')\mathcal{R}(\hat{\gamma}_{\ell})^{1/2}
+\bar{\sigma}\mathcal{R}(\hat{\alpha}_{\ell})^{1/2}
+\frac{\sigma_{\textsc{mis}}}{n^{1/2}} 
\right\}.
    $$
\end{proposition}

\begin{proof}
    I proceed in steps.
    \begin{enumerate}
        \item Decomposition. By Lemma~\ref{lemma:Taylor} and the triangle inequality, write
    \begin{align*}
    |\hat{\theta}-\theta_{\textsc{mis}}|
    &\leq |\hat{\theta}-\bar{\theta}|+|\bar{\theta}-\theta_{\textsc{mis}}|
    \leq\frac{1}{L} \sum_{\ell=1}^L  \left| \hat{\theta}_{\ell}-\bar{\theta}_{\ell}\right|+|\bar{\theta}-\theta_{\textsc{mis}}| 
    \leq \frac{1}{L} \sum_{\ell=1}^L \sum_{j=1}^3 \Delta_{j\ell}^{1/2}+\Delta.
    \end{align*} 
        \item Union bound. 
    Define the event
    $
    \mathcal{E}=(\Delta \leq t)
    $. By Lemma~\ref{lemma:oracle}, $\mathbb{P}(\mathcal{E}^c)\leq \frac{\epsilon}{2}$.
        Define the events
$$
\mathcal{E}'_{\ell}=(\Delta_{1\ell}\leq t_1, \Delta_{2\ell}\leq t_2, \Delta_{3\ell}\leq t_3),\quad \mathcal{E}'=\cap_{\ell=1}^L \mathcal{E}'_{\ell},\quad (\mathcal{E}')^c=\cup_{\ell=1}^L (\mathcal{E}_{\ell}')^c.
$$
Hence by the union bound and Lemma~\ref{lemma:resid2},
$$
\mathbb{P}\{(\mathcal{E}')^c\}\leq \sum_{\ell=1}^L \mathbb{P}\{(\mathcal{E}_{\ell}')^c\} \leq L\frac{\epsilon}{2L}=\frac{\epsilon}{2}.
$$
\item Collecting results. Therefore with probability $1-\epsilon$,
\begin{align*}
    &|\hat{\theta}-\theta_{\textsc{mis}}|  \leq \frac{1}{L} \sum_{\ell=1}^L \sum_{j=1}^3 \Delta_{j\ell}^{1/2}+\Delta  
\leq \frac{1}{L} \sum_{\ell=1}^L \sum_{j=1}^3 t_j^{1/2}+t 
=\sum_{j=1}^3 t_j^{1/2}+t \\
&=\left\{\frac{6L}{\epsilon}\bar{M}\mathcal{R}(\hat{\gamma}_{\ell})\right\}^{1/2}
+\left\{\frac{6L}{\epsilon}(\bar{\alpha}')^2\mathcal{R}(\hat{\gamma}_{\ell})\right\}^{1/2}
+\left\{\frac{6L}{\epsilon}\bar{\sigma}^2\mathcal{R}(\hat{\alpha}_{\ell})\right\}^{1/2}
+\left(\frac{2}{\epsilon}\right)^{1/2}\frac{\sigma_{\textsc{mis}}}{n^{1/2}} \\
&\leq \left(\frac{6L}{\epsilon}\right)^{1/2}\left\{
\bar{M}^{1/2}\mathcal{R}(\hat{\gamma}_{\ell})^{1/2}
+\bar{\alpha}'\mathcal{R}(\hat{\gamma}_{\ell})^{1/2}
+\bar{\sigma}\mathcal{R}(\hat{\alpha}_{\ell})^{1/2}
+\frac{\sigma_{\textsc{mis}}}{n^{1/2}} 
\right\}.
\end{align*}
    \end{enumerate}
\end{proof}

\begin{proof}[Proof of Proposition~\ref{prop:mis}]
    Suppose $\gamma_{\mathcal{G}}=\gamma_0$. Then by the law of iterated expectations,
    $$
    \theta_{\textsc{mis}}=\mathbb{E} [m(W,\gamma_0)+\alpha_{\mathcal{A}}(W)\{Y-\gamma_0(W)\}]=\mathbb{E}\{m(W,\gamma_0)\}=\theta_0
    $$
    and $\sigma_{\textsc{mis}}^2=var\{\psi(W,\theta_0,\gamma_0,\alpha_{\mathcal{A}})\}$.
    Suppose instead that $\alpha_{\mathcal{A}}=\alpha_0$. Then by Riesz representation and the law of iterated expectations,
     \begin{align*}
    \theta_{\textsc{mis}}&=\mathbb{E} [m(W,\gamma_{\mathcal{G}})+\alpha_0(W)\{Y-\gamma_{\mathcal{G}}(W)\}]=\mathbb{E}\{\alpha_0(W)Y\}\\
    &=\mathbb{E}\{\alpha_0(W)\gamma_0(W)\}=\mathbb{E}\{m(W,\gamma_0)\}=\theta_0
      \end{align*}
    and $\sigma_{\textsc{mis}}^2=var\{\psi(W,\theta_0,\gamma_{\mathcal{G}},\alpha_0)\}$.
    Finally appeal to Proposition~\ref{prop:non}.
\end{proof}
\section{Deferred proofs from Sections~\ref{sec:sketch}}\label{sec:sketch_details}

\subsection{Counterfactual effective dimension}

\begin{proof}[Proof of Lemma~\ref{lemma:norms}]
The result follows from Parseval's identity.
\end{proof}

\begin{proof}[Proof of Lemma~\ref{lemma:prior_new}]
By Assumption~\ref{assumption:cont}, there exists some $\bar{M}<\infty$ such that for all $f\in H$, 
$
\mathbb{E}\{m(W,f)^2\}\leq \bar{M} \mathbb{E}\{f(W)^2\}.
$
By the reproducing property, boundedness of the counterfactual kernel, and the definition of $T^{(m)}$,
\begin{align*}
    \mathbb{E}\{m(W,f)^2\}&=\mathbb{E}\left\{\langle f,\phi^{(m)}(W) \rangle^2_H\right\}=\mathbb{E}\left(\langle f,[\phi^{(m)}(W)\otimes \{\phi^{(m)}(W)\}^*]f \rangle_H\right)\\
    &=\langle f,T^{(m)}f \rangle_H.
\end{align*}
By the reproducing property, boundedness of the actual kernel, and the definition of $T$,
$$
\mathbb{E}\{f(W)^2\}=\mathbb{E}\left\{\langle f,\phi(W) \rangle^2_H\right\}=\mathbb{E}\left[\langle f,\{\phi(W)\otimes \phi(W)^*\}f \rangle_H\right]=\langle f,Tf \rangle_H.
$$
In summary, for all $f\in H$,
$
\langle f,T^{(m)}f \rangle_H \leq \bar{M} \langle f,Tf \rangle_H
$
i.e. $\bar{M} T-T^{(m)}  \succeq  0$ in the sense of Loewner ordering. Therefore by properties of trace,
$$
    \bar{M} \mathcal{N}(\lambda)-\mathcal{N}^{(m)}(\lambda)=Tr[(T+\lambda)^{-1}\{\bar{M} T-T^{(m)}\}] \geq 0.
$$
\end{proof}

\subsection{Concentration in the RKHS}

\begin{lemma}[Concentration; Proposition 2 of \cite{caponnetto2007optimal}]\label{lemma:concentration}
Let $(\Omega,\mathcal{F},p)$ be a probability space. Let $\xi$ be a random variable on $\Omega$ taking value in a real separable Hilbert space $\mathcal{H}$. Assume there exist $\bar{a},\bar{b}>0$ such that 
$
\|\xi(\omega)\|_{\mathcal{H}}\leq \frac{\bar{a}}{2}$ and $\mathbb{E}(\|\xi\|^2_{\mathcal{H}})\leq \bar{b}^2.
$
Then for all $n\in\mathbb{N}$ and $\delta\in(0,1)$,
$$
\mathbb{P}_{(\omega_i)\sim p^{n}}\left\{\left\|\frac{1}{n}\sum_{i=1}^n \xi(\omega_i)-\mathbb{E}(\xi)\right\|_{\mathcal{H}}\leq 2 \ln(2/\delta)\left(\frac{\bar{a}}{n}+\frac{\bar{b}}{\sqrt{n}}\right) \right\}\geq 1-\delta.
$$
\end{lemma}

\begin{proof}[Proof of Proposition~\ref{prop:high}]
The argument for $\mathcal{E}_1$ immediately precedes \cite[eq. 41]{caponnetto2007optimal}. The argument for $\mathcal{E}_2$ is identical to \cite[eq. 43]{caponnetto2007optimal}. I prove a new result for $\mathcal{E}_3$ appealing to Lemmas~\ref{lemma:prior_new} and~\ref{lemma:concentration}, the former of which appears to be new.
 Define
    $$
    \xi_i=(T+\lambda)^{-\frac{1}{2}}[\phi^{(m)}(W_i)-\{\phi(W_i)\otimes \phi(W_i)^*\}\alpha_H].
    $$
    By Lemma~\ref{lemma:foc}, 
    $
    \mathbb{E}(\xi_i)=(T+\lambda)^{-\frac{1}{2}}\{\mu^{(m)}-T\alpha_H\}=0
    $. Towards an application of Lemma~\ref{lemma:concentration}, I analyze $\|\xi\|_H$ and $\mathbb{E}(\|\xi\|^2_H)$.
    \begin{enumerate}
        \item To bound $\|\xi\|_H$, write
      \begin{align*}
          \|\xi_i\|_H
          &\leq\|(T+\lambda)^{-\frac{1}{2}}\|_{op}\|\phi^{(m)}(W_i)-\{\phi(W_i)\otimes \phi(W_i)^*\}\alpha_H\|_H \\
          &\leq \lambda^{-1/2}\{\sqrt{\kappa^{(m)}}+\kappa\|\alpha_H\|_H\}
      \end{align*}
      i.e. $\bar{a}=2\lambda^{-1/2}\{\sqrt{\kappa^{(m)}}+\kappa\|\alpha_H\|_H\}$.
        
        \item To bound $\mathbb{E}(\|\xi\|^2_H)$, write
        $$
    \xi_i=A-B,\quad A=(T+\lambda)^{-\frac{1}{2}}\phi^{(m)}(W_i),\quad B=(T+\lambda)^{-\frac{1}{2}}\alpha_H(W_i)\phi(W_i)
    $$
    where I use the reproducing property to simplify $$\{\phi(W_i)\otimes \phi(W_i)^*\}\alpha_H=\phi(W_i) \langle \phi(W_i),\alpha_H\rangle_H=\alpha_H(W_i)\phi(W_i).$$
    Hence
    $
    \|\xi_i\|^2_H\leq 2\|A\|_H^2+2\|B\|_H^2.
    $
    Focusing on the former term, by properties of trace,
    \begin{align*}
        \|A\|_H^2
        &=\langle (T+\lambda)^{-\frac{1}{2}}\phi^{(m)}(W_i),(T+\lambda)^{-\frac{1}{2}}\phi^{(m)}(W_i) \rangle_H \\
        &=\langle (T+\lambda)^{-1}\phi^{(m)}(W_i),\phi^{(m)}(W_i) \rangle_H \\
        &=Tr\{\langle (T+\lambda)^{-1}\phi^{(m)}(W_i),\phi^{(m)}(W_i) \rangle_H\} \\
        &=Tr((T+\lambda)^{-1} [\phi^{(m)}(W_i) \otimes \{\phi^{(m)}(W_i)\}^*]).
    \end{align*}
    Therefore by linearity of trace, the definition of $T^{(m)}$, and Lemma~\ref{lemma:prior_new}, 
    $$
    \mathbb{E}(\|A\|^2_H)=\mathcal{N}^{(m)}(\lambda)\leq \bar{M} \mathcal{N}(\lambda).
    $$
    Focusing on the latter term, by properties of trace,
     \begin{align*}
        \|B\|_H^2
        &=\langle (T+\lambda)^{-\frac{1}{2}}\alpha_H(W_i)\phi(W_i),(T+\lambda)^{-\frac{1}{2}}\alpha_H(W_i)\phi(W_i) \rangle_H \\
        &=\langle (T+\lambda)^{-1}\alpha_H(W_i)\phi(W_i),\alpha_H(W_i)\phi(W_i) \rangle_H \\
         &=Tr\{\langle (T+\lambda)^{-1}\alpha_H(W_i)\phi(W_i),\alpha_H(W_i)\phi(W_i) \rangle_H\} \\
         &=\{\alpha_H(W_i)\}^2 Tr[(T+\lambda)^{-1} \{\phi(W_i) \otimes \phi(W_i)^*\}].
    \end{align*}
    By the reproducing property, Cauchy-Schwartz inequality, and boundedness of the kernel,
    $$|\alpha_H(W_i)|=\langle \alpha_H,\phi(W_i)\rangle_H \leq \|\alpha_H\|_H\cdot \|\phi(W_i)\|_H\leq \sqrt{\kappa}\|\alpha_H\|_H.$$
    Therefore by linearity of trace and the definition of $T$,
    $$
    \mathbb{E}(\|B\|_H^2) \leq \kappa \|\alpha_H\|^2_H   \mathcal{N}(\lambda).
    $$
   In summary,
    $$
    \mathbb{E}(\|\xi\|^2_H)\leq 2(\bar{M}+\kappa \|\alpha_H\|^2_H  ) \cdot \mathcal{N}(\lambda),
    $$
    i.e. $\bar{b}=\sqrt{2(\bar{M}+\kappa \|\alpha_H\|^2_H)} \sqrt{\mathcal{N}(\lambda)}$. 
    
    \item Finally I appeal to Lemma~\ref{lemma:concentration}. With probability $1-\delta/3$,
    \begin{align*}
    &\left\|(T+\lambda)^{-\frac{1}{2}}(\hat{\mu}^{(m)}-\hat{T}\alpha_H)\right\|_H
    \leq 2 \ln(6/\delta)\left(\frac{\bar{a}}{n}+\frac{\bar{b}}{\sqrt{n}}\right) \\
    &= 2 \ln(6/\delta)\left\{\frac{2(\sqrt{\kappa^{(m)}}+\kappa\|\alpha_H\|_H)}{n\sqrt{\lambda}}+\frac{\sqrt{2(\bar{M}+\kappa \|\alpha_H\|^2_H)}\sqrt{\mathcal{N}(\lambda)}}{\sqrt{n}}\right\}
    \\
    &\leq2\ln(6/\delta)\left\{\frac{1}{n}\sqrt{\Upsilon^2\frac{\kappa'}{\lambda}}+\sqrt{\frac{\Sigma^2\mathcal{N}(\lambda)}{n}}\right\}.
    \end{align*}
    \end{enumerate}
\end{proof}

\subsection{Nonasymptotic bound}

\begin{proof}[Proof of Proposition~\ref{prop:abstract}]
I follow the structure of \cite[Proof of Theorem 4]{caponnetto2007optimal}, appealing to Proposition~\ref{prop:high} and the union bound.
\begin{enumerate}
    \item By a decomposition that mirrors \cite[eq. 36]{caponnetto2007optimal},
$$
\|T^{1/2}(\tilde{\alpha}-\alpha_H)\|^2_H\leq 3\left[\mathcal{A}(\lambda)+\mathcal{S}_1\{\lambda,(W_i)\}+\mathcal{S}_2\{\lambda,(W_i)\}\right].
$$
The middle and final terms depend on data $(W_i)$, and the work of this proposition is to bound them. They are 
\begin{align*}
    \mathcal{S}_1\{\lambda,(W_i)\}&=\left\|T^{\frac{1}{2}}(\hat{T}+\lambda)^{-1}\{\hat{\mu}^{(m)}-\hat{T}\alpha_H\}\right\|^2_H; \\ 
    \mathcal{S}_2\{\lambda,(W_i)\}&=\left\|T^{\frac{1}{2}}(\hat{T}+\lambda)^{-1}(T-\hat{T})(\alpha_{\lambda}-\alpha_H)\right\|^2_H.
\end{align*}
    \item Consider the final term. Write
$$
\mathcal{S}_2\{\lambda,(W_i)\}\leq\|T^{\frac{1}{2}}(\hat{T}+\lambda)^{-1}\|^2_{op}\|(T-\hat{T})(\alpha_{\lambda}-\alpha_H)\|^2_H.
$$
By \cite[eq. 39]{caponnetto2007optimal}, under $\mathcal{E}_1$, 
    $
    \|T^{\frac{1}{2}}(\hat{T}+\lambda)^{-1}\|_{op} \leq  \lambda^{-1/2}.
    $
    The event 
$\mathcal{E}_2$ directly guarantees a bound on the other factor. In summary, under $\mathcal{E}_1$ and $\mathcal{E}_2$,
\begin{align*}
    \mathcal{S}_2\{\lambda,(W_i)\}&
    \leq \frac{1}{\lambda}\cdot   \{2\ln(6/\delta)\}^2\cdot2\left[\left\{\frac{2\kappa\sqrt{\mathcal{B}(\lambda)}}{n}\right\}^2+\left\{\sqrt{\frac{\kappa\mathcal{A}(\lambda)}{n}}\right\}^2\right] \\
    &=8\ln^2(6/\delta)\left\{\frac{4\kappa^2\mathcal{B}(\lambda)}{n^2\lambda }+\frac{\kappa\mathcal{A}(\lambda)}{n\lambda}\right\}.
\end{align*}
    
    \item Consider the middle term. Write
$$
\mathcal{S}_1\{\lambda,(W_i)\}\leq \|T^{\frac{1}{2}}(\hat{T}+\lambda)^{-1}(T+\lambda)^{\frac{1}{2}}\|^2_{op}\|(T+\lambda)^{-\frac{1}{2}}\{\hat{\mu}^{(m)}-\hat{T}\alpha_H\}\|^2_H.
$$ By \cite[eq. 47]{caponnetto2007optimal}, under $\mathcal{E}_1$, 
    $
    \left\|T^{\frac{1}{2}}(\hat{T}+\lambda)^{-1}(T+\lambda)^{\frac{1}{2}}\right\|_{op} \leq 2.
    $
    The event $\mathcal{E}_3$ directly guarantees a bound on the other factor, and it is designed to resemble \cite[eq. 48]{caponnetto2007optimal}. Therefore by \cite[eq. 49]{caponnetto2007optimal}, under $\mathcal{E}_1$ and $\mathcal{E}_3$,
$$
\mathcal{S}_1\{\lambda,(W_i)\}\leq 32 \ln^2(6/\delta)\left\{\frac{\kappa' \Upsilon^2}{n^2\lambda}+\frac{\Sigma^2 \mathcal{N}(\lambda)}{n}\right\}. 
$$
\end{enumerate}
\end{proof}

\section{Details for Section~\ref{sec:experiments}}\label{sec:details}

\subsection{Data generating process}

A single observation is a tuple $(Y,D,X)$ where $Y\in\mathbb{R}$, $D\in\{0,1\}$, and $X\in\mathbb{R}^4$. To generate this observation, let $\epsilon_{j}\overset{i.i.d.}{\sim}\textsc{unif}(-0.5,0.5)$. Set $X_1=\epsilon_1$, $X_2=1+2X_1+\epsilon_2$, $X_3=1+2X_2+\epsilon_3$, and $X_4=(X_1-1)^2+\epsilon_4$. Draw treatment as $D\sim \textsc{Bern}\{\Lambda(X_1+X_2+X_3+X_4)/2\}$ where $\Lambda$ is the logistic link. When $D=0$, set $Y=0$; when $D=1$, set $Y+X_1X_2X_3X_4+\nu$ where $\nu\sim\mathcal{N}(0,1/16)$. A random sample consists of 100 independently and identically distributed observations.

\subsection{Tuning}

In debiased machine learning, I use $L=5$ folds. I follow the default tuning for the nonparametric regression estimator $\hat{\gamma}$ from the replication package of \cite{chernozhukov2021simple}. The lasso regression and Riesz representer are fit with a generalization of coordinate wise soft thresholding \cite{chernozhukov2018learning}, the random forest regression with 1000 trees \cite{chernozhukov2018original}, and the neural network regression with a single hidden layer of eight neurons \cite{chernozhukov2018original}.

For the regularization value of KRRR, I used the well known generalized cross validation procedure (GCV) \cite{craven1978smoothing}. GCV is asymptotically optimal \cite{li1986asymptotic}, so it aligns with the theoretical requirements of the $L_2$ rate. 

In the simulations, I consider $c_h\in\{0.25,0.50,1.00\}$. The values $c_h\in\{0.25,0.50\}$ consistently work well. In the application, I fix $c_h=0.50$.

\subsection{Kernel}

I use a binary kernel for treatment $k_{\mathcal{D}}(d,d')=1_{d=d'}$ and Gaussian kernels for covariates, e.g. $k_{\mathcal{X}_1}(x_1,x_1')=\exp\left\{-(x_1-x_1')^2/(2\iota_1^2)\right\}$, combined as a product kernel. This choice satisfies the theoretical requirements of the $L_2$ rate: it is bounded, measurable, and characteristic. 

Each Gaussian kernel has a hyperparameter called the length scale, e.g. $\iota_1$ for $k_{\mathcal{X}_1}(x_1,x_1')$. I follow the median heuristic, setting $\iota_1$ equal to the median interpoint distance of $(X_{1i})$, where the interpoint distance between $X_{1i}$ and $X_{1j}$ is $|X_{1i}-X_{1j}|$. I use the same heuristic for the other Gaussian kernels.

I calculate entries of $K^{(1)}$ from the actual kernel
$$
k(w,w')=\langle\phi(w),\phi(w')\rangle_H=
k_{\mathcal{D}}(d,d')
k_{\mathcal{X}_1}(x_1,x_1')
k_{\mathcal{X}_2}(x_2,x_2')
k_{\mathcal{X}_3}(x_3,x_3')
k_{\mathcal{X}_4}(x_4,x_4'),
$$
entries of $K^{(3)}$ from the counterfactual kernel
\begin{align*}
    k_m(w,w')
    &=\langle\phi^{(m)}(w),\phi(w')\rangle_H \\
    &=
\{k_{\mathcal{D}}(1,d')-k_{\mathcal{D}}(0,d')\}
k_{\mathcal{X}_1}(x_1,x_1')
k_{\mathcal{X}_2}(x_2,x_2')
k_{\mathcal{X}_3}(x_3,x_3')
k_{\mathcal{X}_4}(x_4,x_4'),
\end{align*}
and entries of $K^{(4)}$ from the counterfactual kernel
\begin{align*}
k^{(m)}(w,w')
&=\langle\phi^{(m)}(w),\phi^{(m)}(w')\rangle_H\\
&=
\{k_{\mathcal{D}}(1,1)-k_{\mathcal{D}}(1,0)-k_{\mathcal{D}}(0,1)+k_{\mathcal{D}}(0,0)\}\\
&\quad \cdot k_{\mathcal{X}_1}(x_1,x_1')
k_{\mathcal{X}_2}(x_2,x_2')
k_{\mathcal{X}_3}(x_3,x_3')
k_{\mathcal{X}_4}(x_4,x_4').
\end{align*}
Finally observe that $K^{(2)}=\{K^{(3)}\}^{\top}$.

As discussed in Appendix~\ref{sec:nonparametric}, I estimate $\hat{\alpha}_h(W)=\ell_h(W)\tilde{\alpha}(W)$ in Section~\ref{sec:experiments}. Here, $\tilde{\alpha}$ is the KRRR estimator for the balancing weight $\check{\alpha}_0(D,X)=D\pi_0(X)^{-1}-(1-D)\{1-\pi_0(X)\}^{-1}$, where $\pi_0(X)=\mathbb{E}(D|X)$ is the propensity score. The kernel matrices $K^{(1)},K^{(2)},K^{(3)},K^{(4)}$ defined above are for $\tilde{\alpha}$, and then I calculate $\hat{\alpha}_h(W)=\ell_h(W)\tilde{\alpha}(W)$.

\subsection{Comparison}

For the comparison estimator of \cite{singh2020kernel}, I follow the tuning procedures of that work: the product of Gaussian kernels with length scales tuned by the median heuristic, and regularization tuned by the leave one out cross validation.

\begin{figure}[h]
	\centering
    \subfloat[Lasso]{%
    \includegraphics[height=0.21\textheight]{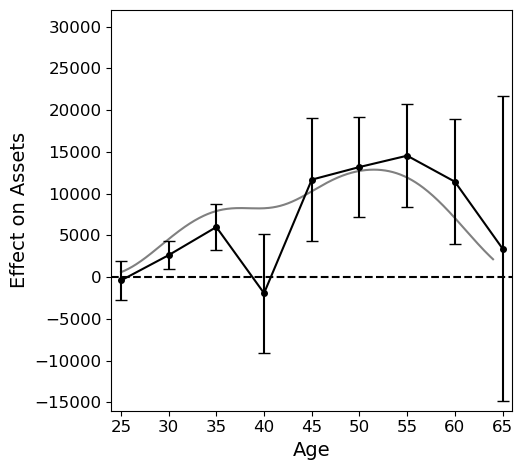}}
    \subfloat[Random forest]{%
    \includegraphics[height=0.21\textheight]{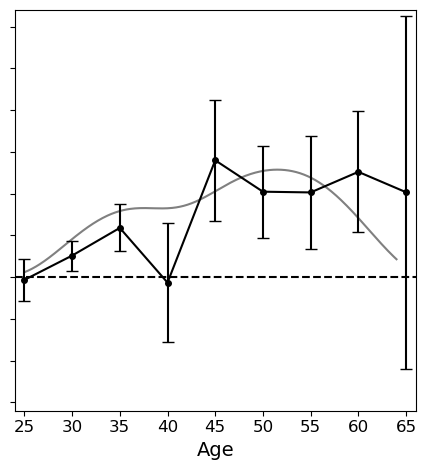}}
    \subfloat[Neural network]{%
    \includegraphics[height=0.21\textheight]{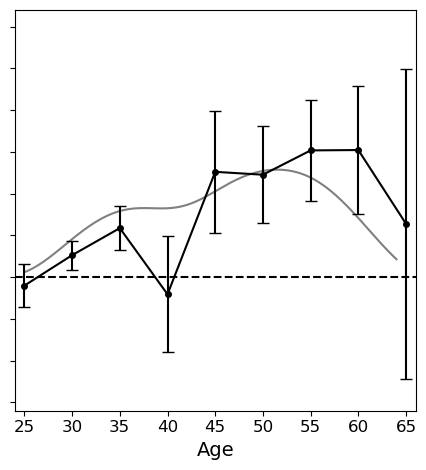}}
    \vspace{-10pt}
    \caption{Heterogeneous treatment effects by age: Low dimensional specification. The point estimates and confidence sets use KRRR.
    The smooth curve is a comparison estimator \cite{singh2020kernel}. 
    }
\label{fig:401_low}
\end{figure}

Finally, I present analogous results using low dimensional specifications. Figure~\ref{fig:401_low} is analogous to Figure~\ref{fig:401_high}, and Table~\ref{tab:low} is analogous to Table~\ref{tab:high}. The findings corroborate those of the high dimensional specifications presented in the main text.

\begin{table*}[h]
\begin{centering}
\caption{Low dimensional coverage simulations}
\label{tab:low}
\setlength{\tabcolsep}{7pt}
\renewcommand{\arraystretch}{0.7}
\resizebox{\textwidth}{!}{%
\begin{tabular}{cccccccccccc}
    \hline
    \multicolumn{3}{c}{} & \multicolumn{3}{c}{Lasso} & \multicolumn{3}{c}{Random Forest} & \multicolumn{3}{c}{Neural Network} \\
    \cmidrule(lr){4-6}\cmidrule(lr){7-9}\cmidrule(lr){10-12}
    & & & Ave. & Ave. & 95\% & Ave. & Ave. & 95\% & Ave. & Ave. & 95\% \\
    $v^*$ & \textsc{cate}($v^*$) & Tuning & Est. & S.E. & Cov. & Est. & S.E. & Cov. & Est. & S.E. & Cov. \\
    \hline
    -0.25 & -0.10 & 0.25 & -0.10 & 0.09 & 99\% & -0.10 & 0.05 & 94\% & -0.12 & 0.07 & 96\% \\
    -0.25 & -0.10 & 0.50 & -0.08 & 0.06 & 99\% & -0.10 & 0.04 & 94\% & -0.11 & 0.05 & 98\% \\
    -0.25 & -0.10 & 1.00 & -0.06 & 0.04 & 97\% & -0.08 & 0.03 & 87\% & -0.09 & 0.03 & 96\% \\
    0.00 & 0.00 & 0.25 & 0.02 & 0.07 & 94\% & 0.00 & 0.03 & 94\% & -0.01 & 0.06 & 96\% \\
    0.00 & 0.00 & 0.50 & 0.03 & 0.06 & 98\% & 0.01 & 0.02 & 93\% & -0.00 & 0.05 & 97\% \\
    0.00 & 0.00 & 1.00 & 0.05 & 0.04 & 92\% & 0.03 & 0.02 & 80\% & 0.02 & 0.04 & 93\% \\
    0.25 & 0.32 & 0.25 & 0.36 & 0.15 & 98\% & 0.39 & 0.19 & 94\% & 0.36 & 0.17 & 94\% \\
    0.25 & 0.32 & 0.50 & 0.37 & 0.11 & 99\% & 0.38 & 0.14 & 95\% & 0.35 & 0.12 & 97\% \\
    0.25 & 0.32 & 1.00 & 0.34 & 0.08 & 99\% & 0.35 & 0.10 & 95\% & 0.33 & 0.10 & 98\% \\
    \hline
\end{tabular}
}
\end{centering}
\end{table*}

\end{document}